\PassOptionsToPackage{table,x11names}{xcolor}
\documentclass[10pt,twocolumn,letterpaper]{article}

\usepackage[pagenumbers]{cvpr}      

\usepackage{amsmath,amsfonts,bm}

\def\rmI{{\mathbf{I}}}

\DeclareMathAlphabet{\mathsfit}{\encodingdefault}{\sfdefault}{m}{sl}
\SetMathAlphabet{\mathsfit}{bold}{\encodingdefault}{\sfdefault}{bx}{n}

\newcommand{\x}{{\boldsymbol x}}

\newcommand{\z}{{\boldsymbol z}}

\newcommand{\epsilonb}{{\boldsymbol \epsilon}}

\newcommand{\Ed}{{\mathbb E}}

\newcommand{\Dc}{{\mathcal D}}
\newcommand{\Ec}{{\mathcal E}}

\newcommand{\Nc}{{\mathcal N}}

\usepackage{amsmath}

\usepackage{algorithm}      
\usepackage{algorithmic}   

\usepackage[utf8]{inputenc} 
\usepackage[T1]{fontenc}    
\usepackage{url}            
\usepackage{booktabs}       
\usepackage{amsfonts}       
\usepackage{nicefrac}       
\usepackage{microtype}      
\usepackage{xcolor}       
\usepackage{xkcdcolors}
\usepackage{times}
\usepackage{epsfig}
\usepackage{graphicx}
\usepackage{placeins}
\usepackage{multirow}
\usepackage{caption}
\usepackage[skip=5pt]{caption}
\usepackage{rotating}
\usepackage{cancel}
\usepackage{setspace}

\usepackage{bm}
\usepackage{bibunits} 

\usepackage{amssymb,amsthm}
\usepackage{amsmath}
\usepackage{graphicx}
\usepackage{wrapfig,lipsum,booktabs}

\usepackage{epsfig}
\usepackage{tikz}
\usetikzlibrary{spy}
\usepackage{algorithm}
\usepackage{mathrsfs}

\usepackage{nicefrac}       
\usepackage{booktabs}       

\usepackage{thmtools,thm-restate}
\usepackage{listings}
\usepackage{lstautogobble}  %
\usepackage{color}          %
\usepackage{zi4}

\usepackage{indentfirst}

\usepackage{makecell}
\usepackage{tabularx}
\usepackage{animate}
\usepackage{capt-of}

\usepackage{colortbl} 
\usepackage[table]{xcolor}

\definecolor{cvprblue}{rgb}{0.21,0.49,0.74}
\usepackage[pagebackref,breaklinks,colorlinks,allcolors=cvprblue]{hyperref}
\usepackage{amsmath}

\def\paperID{15820} 
\def\confName{CVPR}
\def\confYear{2025}

\title{VideoGuide: Improving Video Diffusion Models without Training \\ Through a Teacher's Guide}

\author{Dohun Lee$^*$, Bryan S Kim$^*$, Geon Yeong Park, Jong Chul Ye \\
Kim Jaechul Graduate School of AI, KAIST\\
{\footnotesize $^*$: Equal Contribution}\\
\tt\small{\{leedh7, bryanswkim, pky3436, jong.ye\}@kaist.ac.kr}
}

\graphicspath{{figs/}}

\begin{document}

\twocolumn[{%
\renewcommand\twocolumn[1][]{#1}%
\maketitle
\vspace{-0.7cm}
\begin{center}
    \newcommand{\numColumns}{4}
    \newcommand{\columnSpacing}{0.1cm}
    \begin{tabularx}{\textwidth}{XXXX}
        \centering \small \textbf{Base} & 
        \centering \small \textbf{Ours} & 
        \centering \small \textbf{Base} &
        \centering \small \textbf{Ours}
    \end{tabularx}
    \begin{tabular}{
        @{}
        p{\dimexpr(\textwidth-\columnSpacing*(\numColumns-1))/\numColumns} @{\hspace{\columnSpacing}}
        p{\dimexpr(\textwidth-\columnSpacing*(\numColumns-1))/\numColumns} @{\hspace{\columnSpacing}}
        p{\dimexpr(\textwidth-\columnSpacing*(\numColumns-1))/\numColumns} @{\hspace{\columnSpacing}}
        p{\dimexpr(\textwidth-\columnSpacing*(\numColumns-1))/\numColumns} @{}
    }
        \animategraphics[loop, width=\linewidth]{8}{videos/1_1/}{00}{15} &
        \animategraphics[loop, width=\linewidth]{8}{videos/1_2/}{00}{15} &
        \animategraphics[loop, width=\linewidth]{8}{videos/1_3/}{00}{15} &
        \animategraphics[loop, width=\linewidth]{8}{videos/1_4/}{00}{15}
    \end{tabular}
    \begin{tabularx}{\textwidth}{XX}
        \centering \footnotesize {\fontfamily{put}\selectfont "A drone view of celebration with Christmas tree and fireworks"} & 
        \centering \footnotesize {\fontfamily{put}\selectfont "A boat sailing in the middle of the ocean"}
    \end{tabularx}

    \vspace{0.5em}
    \begin{tabularx}{\textwidth}{XXXX}
        \centering \small \textbf{Base} & 
        \centering \small \textbf{Ours} & 
        \centering \small \textbf{Base} &
        \centering \small \textbf{Ours}
    \end{tabularx}
    \renewcommand{\numColumns}{4}
    \renewcommand{\columnSpacing}{0.25em}
    \begin{tabular}{
        @{}
        p{\dimexpr(\textwidth-\columnSpacing*(\numColumns-1))/\numColumns} @{\hspace{\columnSpacing}}
        p{\dimexpr(\textwidth-\columnSpacing*(\numColumns-1))/\numColumns} @{\hspace{\columnSpacing}}
        p{\dimexpr(\textwidth-\columnSpacing*(\numColumns-1))/\numColumns} @{\hspace{\columnSpacing}}
        p{\dimexpr(\textwidth-\columnSpacing*(\numColumns-1))/\numColumns} @{}
    }
        \animategraphics[loop, width=\linewidth]{8}{videos/2_1/}{00}{15} &
        \animategraphics[loop, width=\linewidth]{8}{videos/2_2/}{00}{15} &
        \animategraphics[loop, width=\linewidth]{8}{videos/2_3/}{00}{15} &
        \animategraphics[loop, width=\linewidth]{8}{videos/2_4/}{00}{15}
    \end{tabular}
    \begin{tabularx}{\textwidth}{XX}
        \centering \footnotesize {\fontfamily{put}\selectfont "Slow motion footage of a racing car"} & 
        \centering \footnotesize {\fontfamily{put}\selectfont "A dog drinking water"}
    \end{tabularx}
    
    \vspace*{0.2cm}
    \captionof{figure}{
    \textbf{VideoGuide} is a novel framework for improving temporal consistency while preserving imaging quality, enabling high-quality video generation for diverse text prompts. By applying VideoGuide to underperforming base models, we can significantly improve temporal consistency with no additional training or fine-tuning. \textit{Best viewed with Acrobat Reader. Click each image to play the video clip.}}
    \label{fig:main}
\end{center}
}]

\begin{abstract}    
Text-to-image (T2I) diffusion models have revolutionized visual content creation, but extending these capabilities to text-to-video (T2V) generation remains a challenge, particularly in preserving temporal consistency. Existing methods that aim to improve consistency often cause trade-offs such as reduced imaging quality and impractical computational time. To address these issues we introduce VideoGuide, a novel framework that enhances the temporal consistency of pretrained T2V models without the need for additional training or fine-tuning. Instead, VideoGuide leverages {\em any} pretrained video diffusion model (VDM) or itself as a guide during the early stages of inference, improving temporal quality by interpolating the guiding model’s denoised samples into the sampling model's denoising process. The proposed method brings about significant improvement in temporal consistency and image fidelity, providing a cost-effective and practical solution that synergizes the strengths of various video diffusion models. Furthermore, we demonstrate prior distillation, revealing that base models can achieve enhanced text coherence by utilizing the superior data prior of the guiding model through the proposed method. Project Page: \url{https://dohunlee1.github.io/videoguide.github.io/}
\end{abstract}

\section{Introduction}
\label{sec:intro}

Text-to-image (T2I) diffusion models have made significant advances in visual generation, enabling user interactive image generation with enriched text descriptions. Now the AI community is looking deeper into the potential of T2I diffusion models, exploring their application to the higher dimensional field of video generation. Text-to-video (T2V) diffusion models aim to extend the capabilities of their image-based counterparts by generating coherent video sequences from text descriptions, handling both spatial and temporal dimensions simultaneously.

Text-to-video (T2V) diffusion models often face a challenging trade-off between temporal consistency and image quality, where improvements in one frequently degrade the other. This compromise results in diminished perceived quality and negatively impacts downstream tasks, such as T2V personalization. Although recent works~\cite{wu2023freeinit, chen2024unictrl} have attempted to address aspects of temporal quality, they often do so at the expense of either visual fidelity or inference speed. In this work, we address the clear need for a robust approach that refines the temporal stability of pretrained T2V models without sacrificing image quality. We propose a novel framework that greatly improves the quality of generated samples without requiring any training or fine-tuning.

Specifically, we introduce VideoGuide, a general framework that uses any pretrained \textit{video} diffusion model as a \textit{guide} during early steps of reverse diffusion sampling. The choice of pretrained teacher VDM is flexible; it can be freely selected from any existing VDMs, or even be the model itself. In any case, the VDM that acts as the guide provides a consistent video trajectory by proceeding in its own denoising for a small number of steps. The teacher model's denoised sample is then interpolated with the original denoising process to guide the sample towards a direction with better temporal quality. Through interpolation, the student VDM is able to follow the temporal consistency of the teacher VDM to produce samples of enhanced quality. Such interpolation only needs to be involved in the first few steps of inference, but is strong enough to guide the entire denoising process towards more desirable results.

VideoGuide is a versatile framework that allows any pretrained video diffusion model to be used for distillation in a plug-and-play fashion. By integrating a high-performing VDM as a video guide, our framework elevates underperforming VDMs to achieve state-of-the-art quality, which is particularly useful when the student model possesses unique traits unavailable for the teacher model. Additionally, we find that interpolating the teacher model's denoised sample provides the student model with an enhanced noise prior, guiding it to generate samples previously unattainable.

In particular, we show two representative cases of how VideoGuide can be applied to combine the best of both worlds: unique functions provided by the student model and high temporal stability provided by the teacher model. In AnimateDiff~\cite{guo2023animatediff}, a motion module is trained that can be interleaved into any pretrained T2I model. The scheme works for any personalized image diffusion model and grants easy application of controllable and extensible modules~\cite{zhang2023adding, guo2023sparsectrl}, but not without consequences. Specifically, fixing the T2I weights limits interaction between the temporal module and generated spatial features, hence harming temporal consistency. Applying VideoGuide with an open-source state-of-the-art model without personalization capability~\cite{chen2024videocrafter2}
as the teacher model, we can greatly enhance the temporal quality of AnimateDiff. Thus, personalization and controllability is provided by the student model, while temporal consistency is refined by the teacher model. Likewise, LaVie~\cite{wang2023lavie} is a multifaceted T2V model that offers various functions including interpolation and super-resolution in a cascaded generation framework, but shows substandard temporal consistency. Using VideoGuide, we can upgrade its temporal consistency with an external model while maintaining its multiple functions. 

The synergistic effects that our framework can bring are not limited to these two cases but are, in fact, boundless. As powerful video diffusion models emerge, existing models will not become obsolete but actually improve through the guidance our method provides. Moreover, as VideoGuide can be applied solely during inference time, these benefits can be enjoyed with no cost at all. Our contributions can be summarized as follows: 
\begin{enumerate}
\item We propose VideoGuide, a novel framework for enhancing temporal consistency and motion smoothness while maintaining the imaging quality of the original VDM.
\item We show how \textit{any} existing VDM can be incorporated into our framework, enabling boosted performance of inadequate models along with newfound synergistic effects among models.
\item We provide evidence of prior distillation, in which the informative prior of teacher models can be utilized to create samples of improved text coherency.
\end{enumerate}

\begin{figure*}
    \centering
    \includegraphics[width=.8\linewidth]{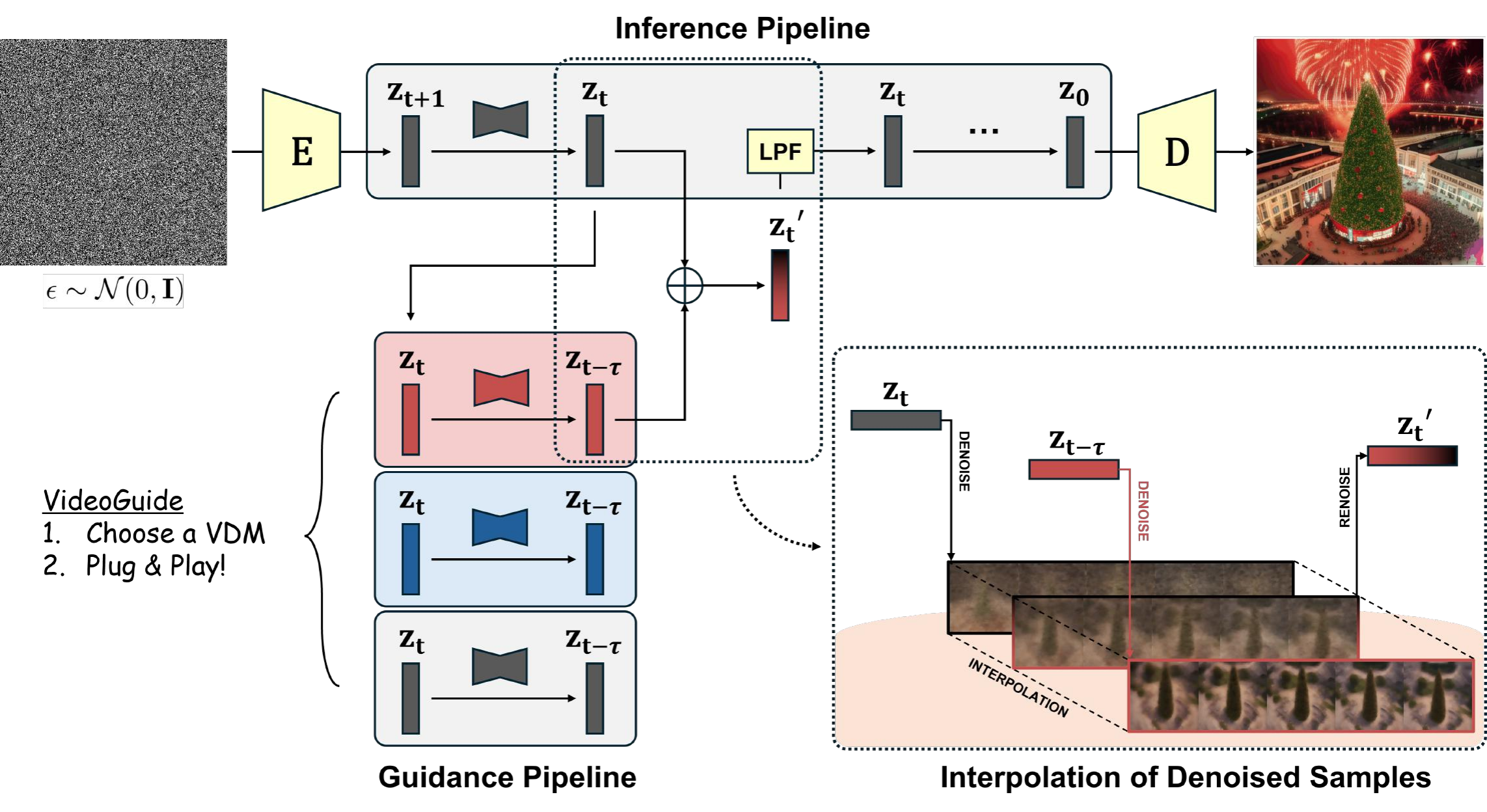}
    \caption{\textbf{Overall Pipeline.} VideoGuide is a framework for enhancing temporal quality without additional training, leveraging the capabilities of any pretrained VDM. Throughout the denoising process of the sampling VDM, the guiding VDM receives an intermediate latent $\z_t$ and provides a temporally consistent sample $\z_{t-\tau}$ by proceeding in its own denoising for a small number of steps $\tau$. The sample $\z_{t-\tau}$ is then denoised and interpolated with the denoised $\z_t$ to produce a fused latent $\z'_{t}$. Such interpolation only needs to take part in the first few steps of inference, and effectively guides samples towards a direction of improved temporal consistency. To further ensure model flexibility in refining high-frequency areas for better image quality, the latent $\z'_{t}$ is passed through a Low-Pass Filter (LPF). Overall, VideoGuide is a straightforward addition to the original pipeline, yet it is powerful enough to significantly enhance temporal consistency without compromising imaging quality or motion smoothness.}
    \label{fig:pipeline}
\end{figure*}

\section{Related Works}
\label{sec:related_work}

\noindent\textbf{The Diffusion Model.}
Diffusion probabilistic models~\citep{ho2020denoising} have achieved great success as generative models. To address the significant computational cost that arises from operating in pixel space, Latent Diffusion Models (LDMs)~\citep{rombach2021highresolution} learn the diffusion process in latent space.
LDMs utilize an encoder-decoder framework where the encoder $\Ec$ and the decoder $\Dc$ are trained together to reconstruct the input data. This training aims to satisfy the relation $\x = \Dc(\z_0) = \Dc(\Ec(\x)) $, where $\z_0$ is the latent representation of the corresponding clean pixel image $\x$. Thus the forward diffusion process in latent space is defined as follows:   \begin{align}\label{eq:forward_ddpm}
    \z_t = \sqrt{\bar\alpha_t} \z_0 + \sqrt{1-\bar\alpha_t}\epsilonb, 
\end{align} where $\bar\alpha_t$ is a pre-determined noise scheduling coefficient, and $\epsilonb \sim \Nc(0, \rmI)$ represents Gaussian noise sampled from a standard normal distribution.
The reverse diffusion process is directed by a score-based neural network, denoted as the diffusion model $\epsilonb_\theta$, which is trained using the denoising score matching framework~\citep{ho2020denoising, song2020score}. The training objective for this model is formulated as follows: \begin{align}\label{eq:training_objective}
\min_\theta \Ed_{t, \epsilonb \sim \Nc(0, \rmI)} ||\epsilonb - \epsilonb_\theta(\z_t, t)||^2_2. 
\end{align}
Following the formulation of DDIM~\citep{song2020denoising}, the reverse deterministic sampling from the posterior distribution $p(\z_{t-1}|\z_t, \z_0)$ is given by:
\begin{align}\label{eq:DDIM_sampling}
\z_{t-1} &= \sqrt{\bar\alpha_{t-1}} \z_{0|t} + \sqrt{1-\bar\alpha_{t-1}}\epsilonb_{\theta}(\z_t,t)
\end{align} 
\begin{align}
\label{eq:tweedie_formula}
\z_{0|t} &= \frac{\z_t - \sqrt{1-\bar{\alpha}_t}\epsilonb_\theta(\z_t, t)}{\sqrt{\bar{\alpha}_t}}
\end{align} 
where the denoised sample at timestep $t$, denoted as $\z_{0|t}$, can be obtained using Tweedie's formula~\cite{efron2011tweedie}.

\noindent\textbf{Guidance as Optimization Problems.}
Applying guidance for diffusion models during sampling time assists diffusion models in exploring the latent space with fidelity to the desired manifold, yielding samples that are tailored to specific criteria. Specifically, guidance can be viewed as addressing the optimization problem $\min_{z \in M} \ell(\z)$ for a given loss function $\ell(\z)$, where $M$ represents the clean data manifold. 

 A direct approach can be seen in the context of diffusion model-based inverse problem solvers~\cite{chung2022improving, chung2023diffusion, chung2024decomposed}. In particular, diffusion posterior sampling (DPS)~\cite{chung2023diffusion} defines the loss function as the manifold-constrained gradient (MCG)~\cite{chung2022improving} of a noisy sample $\z_t \in M_t$. The loss function $\nabla_{\z_t}^{mcg}\ell(\z_t) = \nabla_{\z_t}\ell(\z_{0|t})$ is then incorporated into Eq.~\eqref{eq:DDIM_sampling} as follows:
\begin{equation}
    \begin{aligned}
        \z_{t-1} &= \sqrt{\bar{\alpha}_{t-1}}(\z_{0|t} - \gamma_t\nabla_{\z_t}\ell(\z_{0|t})) \\
        &\quad + \sqrt{1-\bar{\alpha}_{t-1}} \epsilonb_\theta(\z_t, t) 
    \end{aligned}
    \label{eq:dps}
\end{equation}
where $\gamma_t$ denotes step size. To reduce computational overhead, Eq.~\eqref{eq:dps} can be equivalently viewed as follows:
\begin{equation}
    \begin{aligned}
        \z_{t-1} &\approx \sqrt{\bar{\alpha}_{t-1}}(\z_{0|t} - \gamma_t\nabla_{\z_{0|t}}\ell(\z_{0|t})) \\
        &\quad+ \sqrt{1-\bar{\alpha}_{t-1}}\epsilonb_\theta(\z_t, t)
    \end{aligned}
    \label{eq:dds}
\end{equation}
This approach, known as decomposed diffusion sampling (DDS)~\cite{chung2024decomposed}, avoids computing the score Jacobian, aligning with methods from~\cite{poole2022dreamfusion}.

Depending on how the loss function $\ell(\z)$ is defined, it is possible to address various tasks, such as solving inverse problems~\cite{chung2023diffusion} and generating text-conditioned images~\cite{chung2024cfg++}.

In this work, we are the first to address the video consistency problem from the perspective of an optimization problem, introducing a novel function designed to enhance sample quality in the video domain. This function is integrated into the reverse diffusion process, similar to Eq.~\eqref{eq:dds}, to provide a simple yet effective guidance term.

\noindent\textbf{Video Diffusion Model \& Consistent Video Generation.}
The Video Diffusion Model (VDM), originally proposed in~\cite{ho2022video}, operates the diffusion process in the video domain. Similar to LDMs, many recent VDMs~\citep{xing2023dynamicrafter, chen2023videocrafter1,he2022lvdm} are trained in the latent space to reduce computational cost. In Latent VDMs (LVDMs), a temporal layer is incorporated to facilitate frame interaction along the temporal axis during training. By modifying $\z_t$ to $\z_t^{1:N}$ in Eqs. \eqref{eq:forward_ddpm}-\eqref{eq:dds}, the diffusion model can be extended to the video domain. For simplicity, we will use the notation $\z_t$ to represent the latent for video generation instead of $\z_t^{1:N}$.

One of the main challenges in utilizing diffusion models for video generation lies in maintaining temporal consistency. In the video domain, PYoCo~\citep{ge2024preservecorrelationnoiseprior} introduces a carefully designed progressive video noise prior to better leverage image diffusion models for video generation. However, PYoCo primarily focuses on the noise distribution during the training stage and requires extensive fine-tuning on video datasets. Recent studies ~\citep{chen2024unictrl, guo2024i4vgen} aim to enhance video consistency by leveraging techniques such as attention injection and pretrained text-to-image models, but either encounter image degradation issues~\citep{chen2024unictrl} or involve a lengthy pipeline to establish initial states~\citep{guo2024i4vgen}. Other approaches introduce improvements in temporal consistency but are primarily focused on long video generation, making them less suitable for the basic 16-frame scenario~\citep{qiu2023freenoise, reuse2023}.

FreeInit~\citep{wu2023freeinit} addresses the issue of video consistency by iterative refinement of the initial noise. This method aims to resolve the training-inference discrepancy in video diffusion models by reinitializing noise with a spatio-temporal filter, ensuring the refined noise better aligns with the training distribution. While this approach enhances frame-to-frame consistency, it has a significant drawback: repeated iteration leads to the loss of fine details and imaging quality degradation. Additionally, the iterative nature of the method induces high computational costs, prolonging the generation process.


\section{VideoGuide}
\label{sec:main}

In this section, we present \textit{VideoGuide}, a novel guiding framework which improves the video consistency without compromising significant computational costs in contrast to prior works. Our framework is based on a teacher-guided latent optimization objective, which when minimized during the early stage of reverse sampling process, progressively improves the temporal consistency of generated video. We begin by outlining the overall optimization framework.


\subsection{Video Consistency Guidance}
\label{sec:3.1}

An important contribution of this work is revealing that video consistency can be enhanced by recasting guidance as an optimization problem. We introduce a new objective that regularizes the sampling path of the reverse diffusion process to improve the quality of generated video samples:
\begin{align}
    \ell(\z_0;\psi,\epsilonb,t) = ||\epsilonb_{\psi}(\sqrt{\bar{\alpha}_t}\z_0+\sqrt{1-\bar{\alpha}_t}\epsilonb, t) - \epsilonb ||^2_2,
\label{eq:sds_loss}
\end{align}
where $\psi$ conceptually represents a general \textit{teacher} model which can denoise the noisy video latents reasonably well. That said, this regularizer represents an ideal condition that high-quality video samples should satisfy: the ideal in-distribution video samples should be well reconstructed from random perturbations followed by denoising using teacher video diffusion models. 


Then, we can now integrate the optimization step of \eqref{eq:sds_loss} in reverse sampling process as a guidance in terms of denoised video estimates $\z_{0|t}$. This reads a single DDIM iterate as follows:


\begin{equation}
    \begin{aligned}
    \z_{t-1} &= \sqrt{\bar{\alpha}_{t-1}}(\z_{0|t}-\gamma_t\nabla_{z_{0|t}}\ell(\z_{0|t};\psi,\epsilonb, t)) \\ &\quad + \sqrt{1-\bar{\alpha}_{t-1}}\epsilonb_{\theta}(\z_t, t)
    \end{aligned}
    \label{eq:dds_2}
\end{equation}

\noindent Observe that Eq.~\eqref{eq:sds_loss} can be reformulated by swapping $\epsilonb_\psi(\z_t, t)$ into an expression of $\z_t$ and $\z_{0|t}^\psi$, and swapping $\epsilonb$ into an expression of $\z_t$ and $\z_0$ as follows:
\begin{equation}
    \begin{aligned}
    \ell(\z_0;\psi,\epsilonb,t) &= \left\| \frac{\z_t - \sqrt{\bar{\alpha}_t}\z_{0|t}^\psi}{\sqrt{1-\bar{\alpha}_t}} - \frac{\z_t - \sqrt{\bar{\alpha}_t}\z_0}{\sqrt{1-\bar{\alpha}_t}} \right\|^2_2 \\
    &= \frac{\bar{\alpha}_t}{1-\bar{\alpha_t}} \left\|\z_0 - \z_{0|t}^{\psi} \right\|^2_2
    \end{aligned}    
    \label{eq:sds_2}
\end{equation}

\noindent Plugging Eq.~\eqref{eq:sds_2} into Eq.~\eqref{eq:dds_2}, the gradient term dissolves into an interpolation scheme and the update rule can be simplified as follows:
\begin{equation}
    \begin{aligned}
    &\z_{t-1} = \sqrt{\bar{\alpha}_{t-1}}\z' + \sqrt{1-\bar{\alpha}_{t-1}}\epsilonb_\theta(\z_t, t) \\ &\text{where} \quad \z' = \beta \cdot \z_{0|t} + (1-\beta) \cdot \z_{0|t}^\psi
    \end{aligned}
\label{eq:solution}
\end{equation}

\noindent with $\beta = 1 - \frac{2\gamma_t\bar{\alpha}_t}{1-\bar{\alpha}_t}$. $\z_{0|t}^{\psi}$ can be obtained with the renoising process of $\z_{0|t}$ as below:
\begin{equation}
    \begin{aligned}
    &\z_{0|t}^\psi = \frac{\z_t-\sqrt{1-\bar{\alpha}_t}\epsilonb_\psi(\z_t, t)}{\sqrt{\bar{\alpha}_t}} \\
    \text{where} \quad &\z_t = \sqrt{\bar{\alpha}_t}\z_{0|t}+\sqrt{1-\bar{\alpha_t}}\epsilonb, \quad \epsilonb \sim N(0,\rmI)
    \end{aligned}
\label{eq:renoising}
\end{equation}

\noindent Specifically, a key challenge in our approach is the lack of a direct model for $\psi$, which, ideally, would take the form of a consistency model~\cite{song2023consistency}. To address this, we approximate $\psi$ using an iterative reverse sampling method to predict the endpoint of the probability flow ODE (PF-ODE). By performing reverse sampling over multiple steps $\tau$ in the original model $\theta$, we generate samples that serve as a proxy for the PF-ODE endpoint. Consequently, we can redefine the terms as follows:
\begin{align}
    \z_{0|t}^\psi \approx \z_{0|t-\tau}, \quad
    \z' = \beta \cdot \z_{0|t} + (1-\beta) \cdot \z_{0|t-\tau}
\end{align}
This approach offers a cost-effective solution for approximating $\psi$ in video diffusion models, eliminating the need to train a separate consistency model.

To address the increased sampling time caused by interpolating multiple $\tau$ samples at each reverse step, we propose applying a low-frequency filter during the early timesteps of the diffusion process. Recent work~\cite{yu2024scaling} indicates that these initial stages primarily establish low-frequency structures, with high-frequency components contributing minimally to image quality. By introducing a low-frequency filter early in the diffusion trajectory, we can accelerate convergence toward consistent samples without sacrificing quality.

Unlike previous methods~\cite{wu2023freeinit}, which apply the filter only to the initial noise, our approach iteratively applies the filter along the entire diffusion path, ensuring stability throughout the trajectory. Additionally, by incorporating early stopping and leveraging our filter’s ability to rapidly achieve sample consistency, we prevent the image degradation typically observed with prolonged optimization~\cite{kim2023collaborative, hertz2023deltadenoisingscore} of score distillation sampling (SDS) loss~\cite{poole2022dreamfusion}, which is similar to our objective. Specifically, we define our updates using low-pass ($\textit{LPF}_{\gamma}$) and high-pass ($\textit{HPF}_{\gamma}$) filters with a cutoff frequency $\gamma$, streamlining the sampling process while maintaining high quality.
\begin{equation}
   \begin{aligned}
    &\z_{t-1} = \textit{LPF}_\gamma(\z_{t-1}) + \textit{HPF}_{1-\gamma}(\epsilonb) \\
    &\text{where} \quad \epsilonb \sim N(0,\rmI)
    \end{aligned}
\label{eq:LPF}
\end{equation}


\subsection{Guidance with External VDMs}

The assumption of $\psi$ in Sec.~\ref{sec:3.1} holds for any \textit{teacher} model that provides a reliable estimate of the denoised sample. This brings us to realize that $\psi$ does not necessarily have to be approximated by the same base model. It is possible to \textit{plug in} any video diffusion model to approximate $\psi$ and the denoising process would be guided to follow the temporal consistency of the supplemented latent. Here, we demonstrate the steps required for utilizing denoised samples $\z_{0|t-\tau}^{(G)}$ of an external guidance model $G$ to enhance the performance of the base sampling model $S$.

\noindent\textbf{Renoising into the Guidance Domain.}
Different video diffusion models are trained on varying noise schedules and distributions, and matching such discrepancies is a mandatory process. When utilizing a guiding model with conflicting factors, the intermediate latent $\z_t$ of the sampling model must be transformed to align with the noise schedule and distribution of the guiding model. The transformation process can be defined as follows:
\begin{equation}
\begin{aligned}
    &\z_t^{(G)} = \sqrt{\bar{\alpha}_t^{(G)}} \z^{(S)}_{0|t} + \sqrt{1-\bar{\alpha}^{(G)}_t} \epsilonb\\ &\text{where} \quad \epsilonb \sim N(0,\rmI)
\end{aligned}
\label{eq:ext}
\end{equation}
where $(S)$ denotes the components related to the base sampling model and $(G)$ denotes the components related to the external guiding model.  Specifically, $\z^{(S)}_{0|t}$ is the denoised sample from $\z^{(S)}_t$ at timestep $t$, and $\bar{\alpha}_t^{(G)}$ is derived from the noise schedule of the guiding diffusion model. The resulting outcome $\z_t^{(G)}$ can then be denoised with the guiding model for a sufficient number of timesteps $\tau$ up to $\z_{0|t-\tau}^{(G)}$. Thus, the PF-ODE endpoint originally approximated by $\psi$ is now represented as $\z_{0|t-\tau}^{(G)}$, the denoised output of the guiding model $G$.

\noindent\textbf{Interpolation of Denoised Samples.}
Interpolating the denoised samples of the two models \textit{S} and \textit{G} can be expressed as below:
\begin{equation}
\begin{aligned}
    \z_{t-1}^{(S)}&=\sqrt{\bar\alpha_{t-1}}(\beta \cdot \z_{0|t}^{(S)} + (1-\beta)\cdot \z_{0|t-\tau}^{(G)}) \\ &\quad + \sqrt{1-\bar\alpha_{t-1}}\epsilonb_{\theta}^{(S)}(\z_t,t)
\end{aligned}
\label{eq:interpolation_diff}
\end{equation}

\noindent Note that the only difference from Eq. (\ref{eq:solution}) is the introduction of the $\z_{0|t-\tau}^{(G)}$ term, where originally $\z_{0|t-\tau}^{(S)}$ would be used. $\textit{LPF}_\gamma$ can then be used on $\z_{t-1}^{(S)}$ as in Eq. (\ref{eq:LPF}) for replacing high-frequency components:
\begin{equation}
\begin{aligned}
    &\z_{t-1}^{(S)} = \textit{LPF}_\gamma(\z^{(S)}_{t-1}) + \textit{HPF}_{1-\gamma}(\epsilonb) \\  &\text{where} \quad \epsilonb \sim N(0, \rmI)
\end{aligned}  
\end{equation}

\noindent In cases where external video diffusion models (VDMs) are used as guidance, the low-frequency filter serves an additional role in preserving domain fidelity. By controlling high-frequency components and introducing Gaussian noise, it prevents unwanted domain drift toward the guidance model while selectively distilling only its temporal stability. This approach captures the temporal consistency of the guiding diffusion model while preserving unique characteristics, such as those in AnimateDiff, without compromising image quality. Notably, these synergistic effects are achieved without additional training or fine-tuning, enabling users to flexibly employ preferred video diffusion models (VDMs) in a plug-and-play manner.

\subsection{Prior Distillation}
Each video diffusion model spans its own specific data distribution, causing sample generation to be restricted to the data prior the model has been trained on. Thus, if the data prior of a model is substandard, the generation results of the model are also inherently substandard. This is especially noticeable when using personalized text-to-image (T2I) models such as Dreambooth or LoRA in AnimateDiff, in which substandard results that do not align with the given text prompt are frequently observed. Prior work~\citep{ge2024preservecorrelationnoiseprior} elaborates on the importance of data prior for VDMs, but the proposed solution involves extensive fine-tuning, making it impractical for simple use cases. 
On the other hand, VideoGuide comes as a potential solution in such cases, where the interpolation between two models exhibit a form of prior distillation. Through the guidance of a generalized video diffusion model (\textit{e.g.}~\cite{chen2024videocrafter2}) the base sampling model is able to refer to the denoised sample provided by the guidance model, and steer its sampling process towards a relevant outcome. This allows for the effective generation of diverse objects, even while retaining the style of the original data domain. For the case of AnimateDiff, the approach allows for broader customization without the need for directly training the personalized T2I model on a wider range of data. Extensive analysis concerning this issue is provided in Sec. \ref{sec:prior}.

\section{Experiments}
\label{sec:experiments}

\noindent\textbf{Experimental Settings.} In our experiments, we leverage multiple open-source Text-to-Video (T2V) diffusion models to explore the combined strengths of each. For the guiding diffusion model, we choose Videocrafter2~\citep{chen2024videocrafter2} due to its strong performance in temporal consistency, as measured by the VBench~\citep{huang2023vbench} benchmark. For sampling, we employ AnimateDiff~\citep{guo2023animatediff} for flexible personalization of video content, and Lavie~\citep{wang2023lavie} to enhance video quality and increase frame count through super-resolution and interpolation techniques. This integration combines the temporal consistency of the guiding model with the advantages of the sampling model. All experiments were conducted using DDIM with 50 steps for sampling. For our experiments with AnimateDiff, we set the number of interpolation steps $I=5$, $\beta=0.5$, and $\tau=10$, and used the Butterworth filter with a normalized frequency of $0.25$ and a filter order of $n=4$. Additional experimental details are provided in Supplementary Material.

\noindent\textbf{Evaluation Metrics.}
To validate the improvement in video consistency with our proposed method, we evaluate five key metrics: subject consistency, background consistency, imaging quality, motion smoothness, and dynamic degree. For subject consistency evaluation, DINO~\citep{caron2021emerging} feature similarity between frames is measured to assess consistency of the subject's appearance throughout the video. Background consistency is evaluated using CLIP feature similarity between frames to evaluate overall scene consistency. Imaging quality is also a key metric in that maintaining original image quality is essential for generation and enabling customization. Thus we evaluate image quality using the multi-scale image quality transformer (MUSIQ)~\citep{ke2021musiq}, which measures frame-wise low-level distortion such as noise, blur, and over-exposure. To ensure smooth motion, we employ a video interpolation model~\citep{licvpr23amt} to assess consistency of motion across video frames. To compare the magnitude of the motion in the videos, we utilize RAFT~\cite{teed2020raftrecurrentallpairsfield} to estimate the optical flow, and calculate the mean of top 5\% largest magnitude of the flows.

\subsection{Comparison Results}
\begin{figure*}
    \centering
    \includegraphics[width=.95\textwidth]{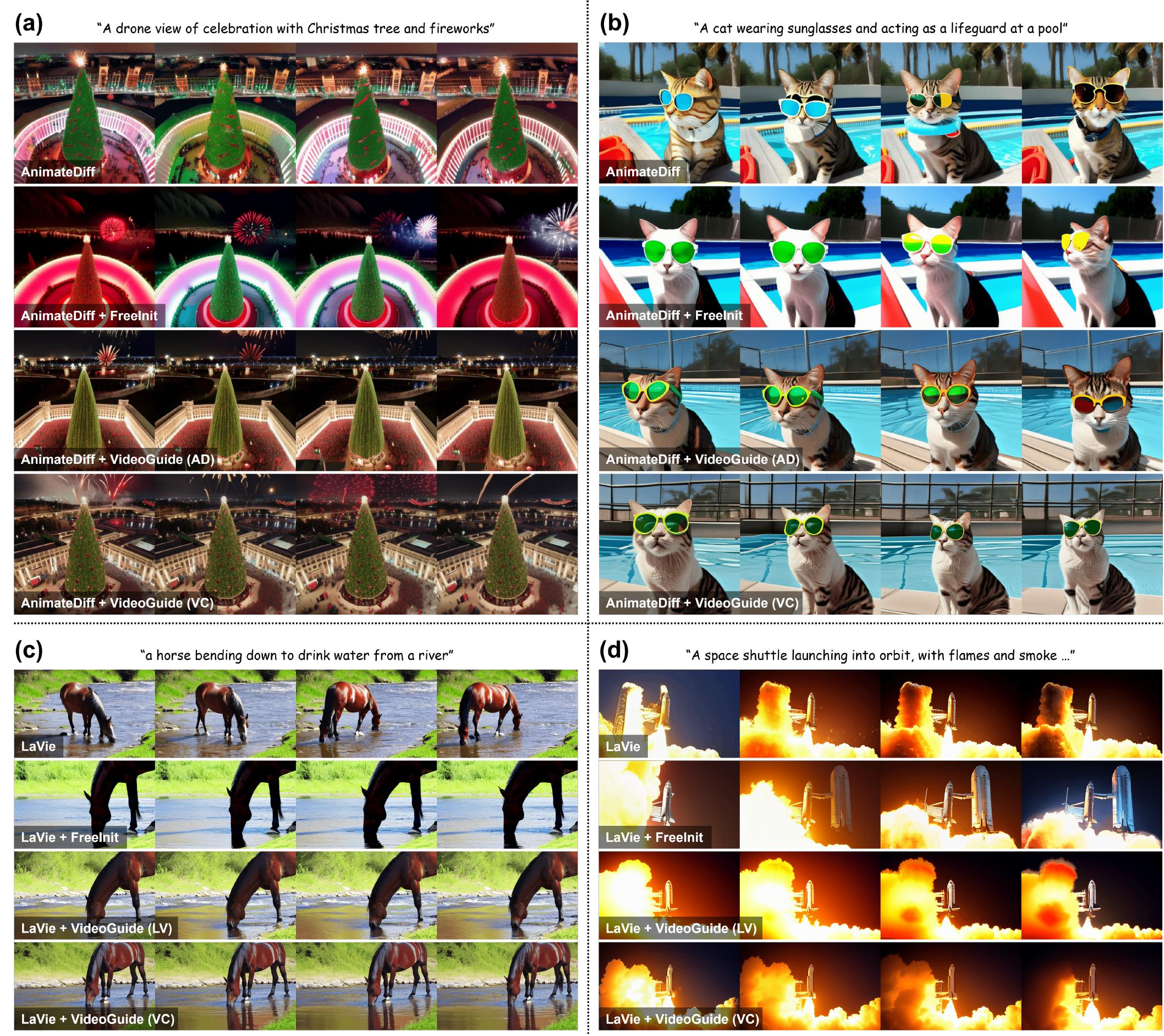}
    \caption{\textbf{Qualitative Comparison.} VideoGuide is applied on various base models for different text prompts.
  For each prompt, frames of generated samples from four different models are displayed: (i) \textbf{Base model} (first row); (ii) \textbf{Base model with FreeInit} (second row); (iii) \textbf{Base model with VideoGuide (self-guided case)} (third row); (iv) \textbf{Base model with VideoGuide (external model-guided case)} (fourth row). AD, VC, LV indicate guidance models of AnimateDiff, VideoCrafter-2.0, LaVie, respectively. Samples for the base model show substandard temporal consistency, especially regarding color fluctuation and subject appearance change. Applying FreeInit improves consistency but introduces degradation in imaging quality, such as smoothing out of textural details. In contrast, applying VideoGuide significantly enhances temporal consistency while preserving imaging quality, both for the self-guided and the external model-guided case.}
    \label{fig:qualitative}
\end{figure*}

Qualitative results for various prompts and base models are shown in Fig. \ref{fig:qualitative}. Samples from the base model show impairment in temporal consistency, such as fluctuation in color or abrupt change in subject appearance. FreeInit~\cite{wu2023freeinit} moderately solves the problem of temporal consistency but at the cost of considerable degradation in imaging quality, such as smoothing out of textural details. In contrast, the proposed VideoGuide significantly enhances temporal consistency without loss of imaging quality. Furthermore, VideoGuide solves sudden frame shifts frequently observed in LaVie by providing smooth frame transitions. Additional qualitative results are included in Supplementary Material.

\begin{table*}
\centering
\resizebox{\textwidth}{!}{%
\fontsize{6}{8}\selectfont
\begin{tabularx}{\textwidth}{l>{\centering\arraybackslash}X>{\centering\arraybackslash}X>{\centering\arraybackslash}X>{\centering\arraybackslash}X>{\centering\arraybackslash}X} 
\toprule
\textbf{Method} & \makecell{\textbf{Subject} \\ \raggedright\textbf{consistency} ($\uparrow$)} & \makecell{\textbf{Background} \\ \textbf{Consistency} ($\uparrow$)} & \makecell{\textbf{Imaging} \\ \textbf{Quality ($\uparrow$)}} &  \makecell{\textbf{Motion} \\ \textbf{Smoothness} ($\uparrow$)} & \makecell{\textbf{Dynamic} \\ \textbf{Degree} ($\uparrow$)} \\

\cmidrule{1-6}
AnimateDiff~\citep{guo2023animatediff} & $0.9183$ & $0.9437$ & $\underline{0.6647}$ & $0.9547$ & $\bm{26.67}$ \\
AnimateDiff + FreeInit~\citep{wu2023freeinit} & $0.9487$ & $\underline{0.9604}$ & $0.6173$ & $0.9705$ & $\underline{19.28}$ \\
AnimateDiff + Ours (with AnimateDiff) & $\underline{0.9520}$ & $0.9600$ & $0.6566$ & $\underline{0.9731}$ & $15.25$ \\
AnimateDiff + Ours (with VideoCrafter2) & $\bm{0.9614}$ & $\bm{0.9664}$ & $\bm{0.6671}$ & $\bm{0.9772}$ & $16.78$ \\

\cmidrule{1-6}
LaVie~\citep{wang2023lavie} & $0.9534$ & $0.9599$ & $0.6750$ & $0.9658$ & $\bm{14.69}$ \\
LaVie + FreeInit~\citep{wu2023freeinit} & $0.9625$ & $\underline{0.9643}$ & $0.6533$ & $\bm{0.9757}$  & $10.69$\\
LaVie + Ours (with Lavie) & $\underline{0.9629}$ & $\bm{0.9652}$ & $\underline{0.6780}$ & $\underline{0.9725}$ & $12.39$ \\
LaVie + Ours (with VideoCrafter2) & $\bm{0.9635}$ & $\underline{0.9643}$ & $\bm{0.6796}$ & $0.9723$ & $\underline{12.63}$ \\

\bottomrule
\end{tabularx}
}
\caption{Quantitative comparison of video generation. \textbf{Bold}: best, \underline{underline}: second best.}
\label{tab:results_video}
\end{table*}

Quantitative comparison shows that our approach consistently outperforms the baseline model, achieving substantial improvements in both subject and background consistency. When using AnimateDiff~\cite{guo2023animatediff} as the baseline, our method delivers the best results across all key metrics, except for dynamic degree, due to an inherent trade-off. Our method greatly enhances temporal consistency with a comparatively small reduction in dynamic degree, ensuring stable temporal coherence without compromising image quality. For LaVie~\cite{wang2023lavie}, our method achieves higher temporal consistency and an increased dynamic degree compared to previous methods, highlighting its ability to enhance temporal stability with minimal impact on dynamic motion.

Incorporating an external model~\cite{chen2024videocrafter2} for guidance enhances performance by achieving higher temporal consistency without compromising dynamic motion, compared to self-guided cases in both AnimateDiff and LaVie. This result indicates that using a higher-performing external model for guidance can lead to superior video quality. Additionally, our method successfully addresses challenges such as sudden visual shifts, as discussed in Supplementary Material.

\begin{table}
\centering
\setlength{\tabcolsep}{0.2em}
\renewcommand{\arraystretch}{1.2} 
\fontsize{16}{16}\selectfont 
\resizebox{0.48\textwidth}{!}{%
\begin{tabularx}{\textwidth}{>{\centering\arraybackslash}X >{\centering\arraybackslash}X >{\centering\arraybackslash}X} 
\toprule
{\textbf{Method}} & \textbf{AnimateDiff} & \textbf{LaVie} \\
\midrule
Baseline & $\bm{11.38}$ & $\bm{6.86}$ \\
FreeInit & $51.98$ & $30.18$ \\
Ours (self-guided) & $\underline{21.68}$ & $\underline{10.01}$ \\
Ours (VC-guided) & $29.73$ & $14.07$ \\ 
\bottomrule
\end{tabularx}
}
\caption{Inference time for video generation($s$). Both the self-guided case and the VideoCrafter2-guided case show significant decrease in inference time compared with previous method~\cite{wu2023freeinit}. \textbf{Bold}: best, \underline{underline}: second best.}
\label{tab:computational_time}
\end{table}

Regarding computational efficiency, iterative initial noise refinement in prior work~\citep{wu2023freeinit} requires performing DDIM sampling over multiple iterations, resulting in high computational cost. In contrast, our method only introduces a small number of additional sampling steps. This difference leads to a significant reduction in inference time, yielding a $\times1.8 \sim \times2.4$ improvement in generation speed for AnimateDiff and a $\times2.1 \sim \times3.0$ improvement for Lavie as shown in Tab. ~\ref{tab:computational_time}.

Our approach achieves these advancements without sacrificing image quality, highlighting its importance for both personalization-focused applications and high-quality video generation. Overall, VideoGuide proves to be essential for optimizing temporal coherence and enhancing overall quality across various model frameworks.

\section{Analysis}
\subsection{Ablation Study}
\label{sec:ablation}

\begin{table}
\centering
\small
\resizebox{1.0\columnwidth}{!}{ 
\begin{tabular}{ccc}
    \begin{tabular}{c c c}
    \toprule
    \multicolumn{3}{c}{\textbf{Interpolation Scale $\beta$}} \\
    \cmidrule(lr){1-3}
    & \textbf{SC} & \textbf{BC} \\
    \midrule
    $\beta=0.9$ & $0.9518$ & $0.9599$ \\
    \addlinespace
    $0.8$ & $0.9546$ & $0.9609$ \\
    \addlinespace
    $0.7$ & $0.9576$ & $0.9628$ \\
    \addlinespace
    $0.6$ & $\underline{0.9605}$ & $\underline{0.9649}$ \\
    \addlinespace
    $0.5$ & $\bm{0.9614}$ & $\bm{0.9664}$ \\
    \bottomrule
    \end{tabular} &

    \begin{tabular}{c c c}
    \toprule
    \multicolumn{3}{c}{\textbf{Interpolation Step Number $I$}} \\
    \cmidrule(lr){1-3}
    & \textbf{SC} & \textbf{BC} \\
    \midrule
    $I=1$ & $0.9524$ & $0.9618$ \\
    \addlinespace
    $2$ & $0.9489$ & $0.9588$ \\
    \addlinespace
    $3$ & $0.9546$ & $0.9612$ \\
    \addlinespace
    $4$ & $\underline{0.9602}$ & $\underline{0.9645}$ \\
    \addlinespace
    $5$ & $\bm{0.9614}$ & $\bm{0.9664}$ \\
    \bottomrule
    \end{tabular} &

    \begin{tabular}{c c c}
    \toprule
    \multicolumn{3}{c}{\textbf{Guidance Step Number $\tau$}} \\
    \cmidrule(lr){1-3}
    & \textbf{SC} & \textbf{BC} \\
    \midrule
    $\tau=1$ & $0.9444$ & $0.9558$ \\
    \addlinespace
    $3$ & $0.9531$ & $0.9611$ \\
    \addlinespace
    $5$ & $0.9582$ & $0.9641$ \\
    \addlinespace
    $7$ & $\underline{0.9611}$ & $\underline{0.9658}$ \\
    \addlinespace
    $10$ & $\bm{0.9614}$ & $\bm{0.9664}$ \\
    \bottomrule
    \end{tabular}
\end{tabular}
}
\caption{Ablation study regarding interpolation scale $\beta$, number of interpolation steps $I$, and number of guidance sampling steps $\tau$. Subject consistency (SC) and background consistency (BC) is compared for various parameters. \textbf{Bold}: best, \underline{underline}: second best.}
\label{tab:ablation_param}
\end{table}

\noindent\textbf{Parameter Selection.}
An analysis is performed to assess how varying parameters of the guiding diffusion model impacts temporal consistency. Specifically, we examine the effects of three factors: interpolation scale $\beta$, number of interpolation steps $I$, number of guidance sampling steps $\tau$.

Our ablation studies prove that all three parameters are closely related to temporal consistency. Decrease in interpolation scale $\beta$ leads to improved subject and background consistency. Note that the minimum value of $\beta$ is constrained to $0.5$ to mitigate the risk of distribution shift. Increasing the number of interpolation steps $I$ also leads to improvement in temporal consistency, which proves that our interpolation scheme is indeed effective. Furthermore, increasing the number of guidance sampling steps $\tau$ enhances consistency, indicating that blending intermediate latents with better-denoised versions enhances overall consistency as expected. Such ablation study highlights the trade-off between consistency improvement and computational efficiency, offering insight into optimal parameter settings for the guiding diffusion model.

\subsection{Prior Distillation}
\label{sec:prior}

\begin{figure}
  \includegraphics[width=1.0\linewidth]{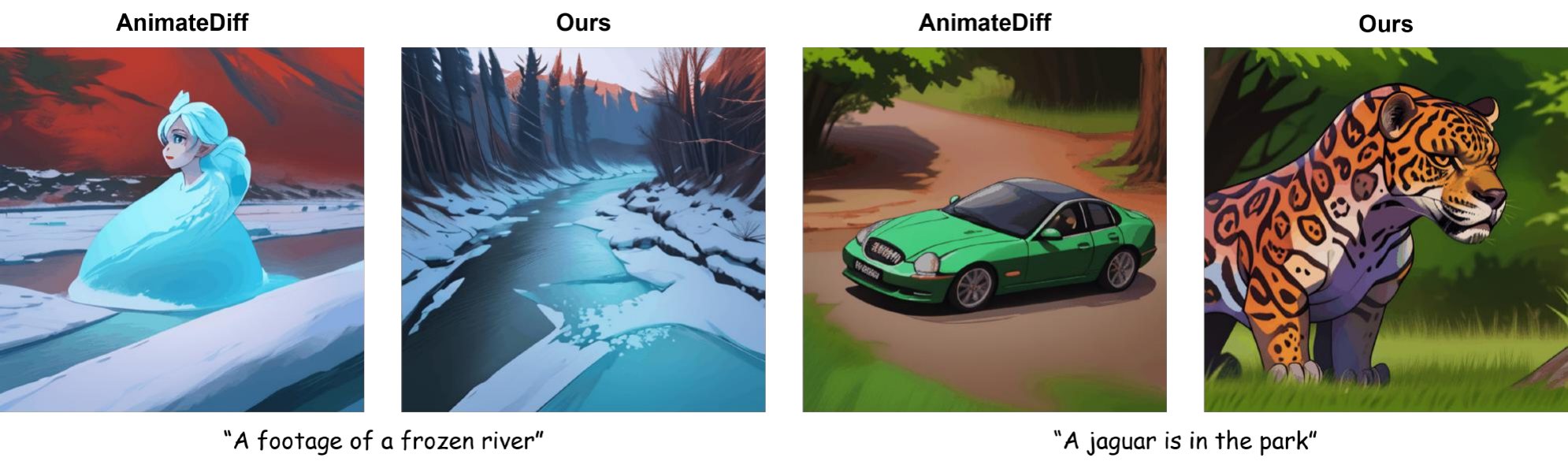}
  \caption{\textbf{Prior Distillation Results.} VideoGuide solves degraded performance regarding text coherency by enabling the utilization of a superior data prior. Example results for certain ambiguous prompts are displayed. For each prompt, the same random seed is shared for both methods. AnimateDiff directs generation of `beetle' and `jaguar' towards car samples due to a substandard data prior. Using VideoGuide, users can distill a superior prior for correct generation.}
  \label{fig:prior_distillation}
\end{figure}

Degraded performance due to a substandard data prior is an issue only solvable through extra training. However VideoGuide provides a workaround to this matter by enabling the utilization of a superior data prior. Fig. \ref{fig:prior_distillation} demonstrates example cases. For all instances, generated samples are guided towards a result of better text coherence while maintaining the style of the original data domain. Additional examples of prior distillation are provided in Supplemental Material.

\section{Conclusion}
\label{sec:conclusion}

In this work, we introduced VideoGuide, a novel and versatile framework that enhances the temporal quality of pretrained text-to-video (T2V) diffusion models without the need for additional training or fine-tuning. Our approach provides temporally consistent samples to intermediate latents during the early stages of the denoising process, guiding the low frequency components of latents towards a direction of high temporal consistency. The samples provided are not confined to the base model; any superior pretrained VDM can be selected for distillation. By doing so, we empower underperforming models with improved motion smoothness and temporal consistency while maintaining their unique traits and strengths, including personalization and controllability. We demonstrate the effectiveness of VideoGuide on various base models, and prove its ability to enhance temporal consistency without sacrifice of imaging quality or motion smoothness compared to prior methods. The potential of VideoGuide extends far beyond the cases discussed, as VideoGuide ensures that even existing models can remain relevant and competitive by leveraging the strengths of superior models. As video diffusion models continue to evolve, new and emerging VDMs will only enhance the pertinence of VideoGuide over time, broadening the scope of VDMs utilizable as a video guide.

{
    \small
    \bibliographystyle{ieeenat_fullname}
    \bibliography{main}
}

\clearpage
\appendix
\setcounter{page}{1}
\maketitlesupplementary

\noindent The supplementary material is organized as follows:
\begin{itemize}
    \item Section \ref{suppl:filter}: Importance of the low-pass filter.
    \item Section \ref{suppl:guidance}: About classifier-free guidance.
    \item Section \ref{suppl:pseudocode}: Pseudocodes for our algorithm.
    \item Section \ref{suppl:experimental-details}: More experimental details.
    \item Section \ref{suppl:user-study}: More quantitative results: user study.
    \item Section \ref{suppl:dit-based}: Application on DiT-based models.
    \item Section \ref{suppl:orthogonal}: Comparison with orthogonal methods.
    \item Section \ref{suppl:limitations}: Limitations.
    \item Section \ref{suppl:qualitative}: More qualitative examples.
\end{itemize}

\section{Importance of Low-Pass Filter}
\label{suppl:filter}

\begin{figure*}[h]
    \centering
    \includegraphics[width=1.0\linewidth]{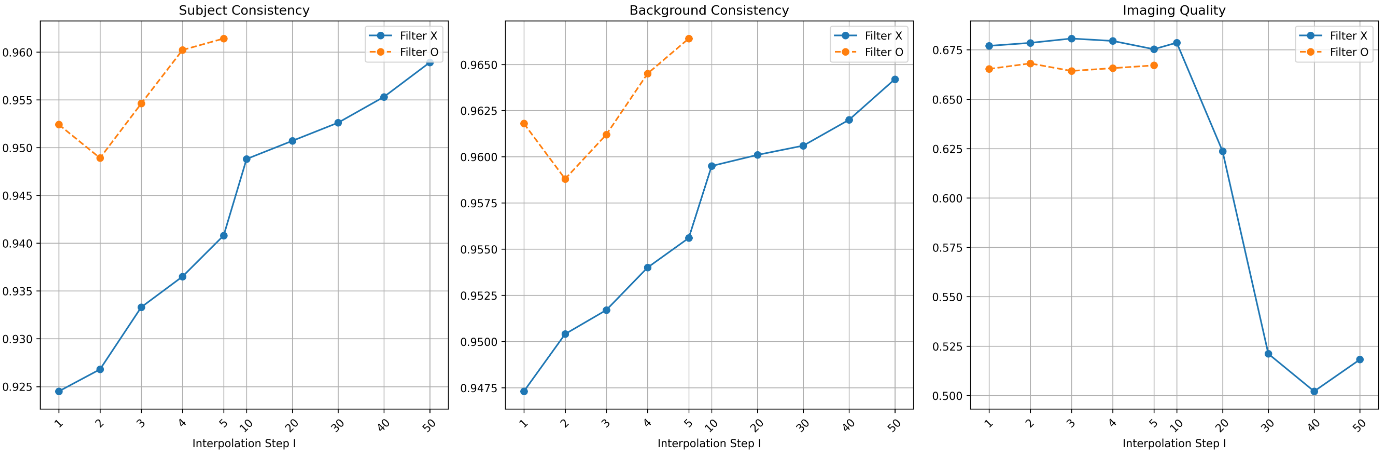}
    \caption{Comparison of Subject Consistency, Background Consistency, and Imaging Quality across interpolation steps ($I$) with and without the application of the low-frequency filter. Results indicate that the low-frequency filter accelerates convergence towards consistency while maintaining imaging quality.}
    \label{fig:filter-graph}
\end{figure*}

To evaluate the role of the low-pass filter in our methodology, we conduct experiments by varying the interpolation step $I$, both with and without the low-pass filter. These experiments are averaged across 800 prompts from the VBench categories for consistent evaluation. We apply the low-pass filter for the initial 5 timesteps, based on the observation that the mid-to-late timesteps in the diffusion process focus on generating mid- and high-frequency details. Replacing these frequencies with random components via the low-pass filter in the mid-to-late timesteps would result in degraded video quality, necessitating the early-timestep limitation. All corresponding results are presented in Fig.~\ref{fig:filter-graph}.

We measure the effects of the filter on Subject Consistency, Background Consistency, and Imaging Quality. Both Subject Consistency and Background Consistency steadily improves as the number of interpolation steps increases, demonstrating the effectiveness of our interpolation scheme in enhancing temporal coherence. Meanwhile, Imaging Quality is maintained up to approximately 10 interpolation steps without the low-pass filter. Beyond this point, a significant drop in quality is observed, indicating that excessive interpolation exacerbates the blurring effects caused by prolonged SDS optimization, as noted earlier in this work.

The improvement in consistency is significantly accelerated when using the low-pass filter. This acceleration is achieved while mitigating the decline in imaging quality typically associated with increased interpolation steps. Furthermore, application of the filter also reduces computational overhead during interpolation. Specifically, the consistency achieved at $I=4$ with the filter is comparable to the consistency achieved at $I=50$ without the filter, offering approximately a 7-fold reduction in inference time. Such results prove the effectiveness of the low-pass filter in balancing consistency improvement, imaging quality preservation, and computational efficiency.

\section{Classifier-Free Guidance}
\label{suppl:guidance}

\begin{figure}[h]
  \centering
  \includegraphics[width=0.8\linewidth]{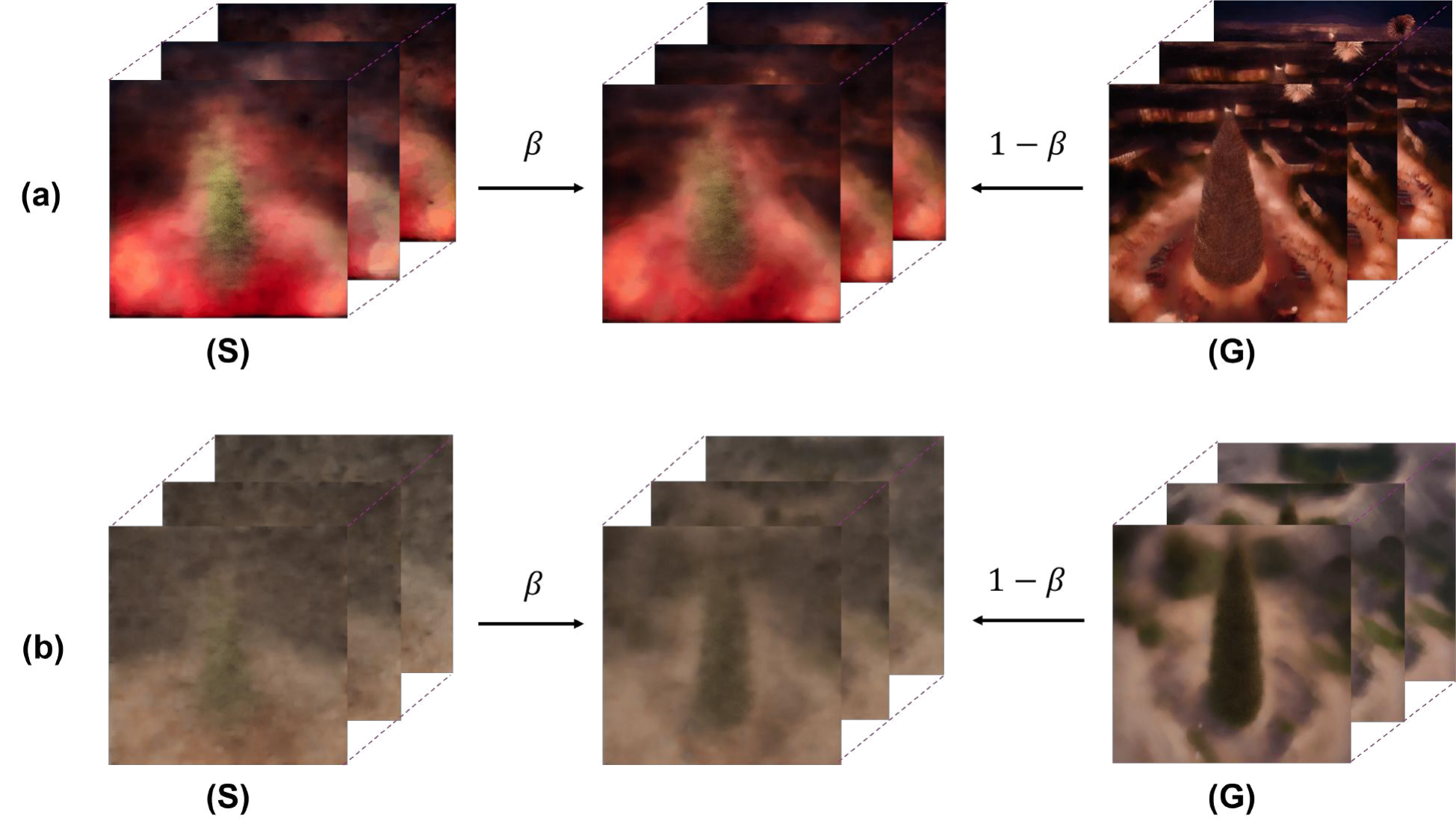}
  \caption{(a) The interpolation process between denoised samples from the sampling model (S) and the guiding model (G) for high guidance scale $w=7.5$ is shown. (b) The interpolation process for low guidance scale $w=0.8$ is shown. Both interpolations are performed at $T=980$ and $\beta=0.7$. Results indicate that with high guidance scale $w$, influence of the guiding diffusion model is significantly reduced due to color saturation.}
\label{fig:ablation_cfg}
\end{figure}

\noindent \textbf{Off-Manifold Behavior of CFG} Recent study~\citep{chung2024cfg++} demonstrates that employing a high Classifier- Free Guidance (CFG)~\citep{ho2021classifierfree} scale ($w > 1.0$) in the early timesteps of diffusion sampling leads to off-manifold behavior. This phenomenon results in denoised samples exhibiting problems such as color saturation and abrupt transitions, which negatively affect the interpolation between samples during these timesteps. We solve this by applying a lower guidance scale $w$ during the early stages of sampling, ensuring smoother interpolation between the denoised samples. As illustrated in Fig.~\ref{fig:ablation_cfg} (a), when using a high CFG scale ($w = 7.5$), the influence of the guiding diffusion model becomes minimal due to significant color saturation, making it difficult for the output of the guiding model to be reflected effectively. In contrast, as illustrated in Fig.~\ref{fig:ablation_cfg} (b), a lower CFG scale ($w = 0.8$) facilitates smoother interpolation between the sampling diffusion model and the guiding diffusion model. 

%


\begin{table}[h]
    \centering
    \small
    \begin{tabular}{lcc}
        \toprule
        \textbf{Configuration} & \textbf{SC ($\uparrow$)} & \textbf{BC ($\uparrow$)} \\
        \midrule
        \rowcolor[gray]{0.9} \multicolumn{3}{l}{\textbf{Base}} \\
        \hspace{1em} + CFG & $0.9183$ & $0.9437$ \\
        \hspace{1em} + CFG++ & $0.9176$ & $0.9435$ \\
        \midrule
        \rowcolor[gray]{0.9} \multicolumn{3}{l}{\textbf{FreeInit}} \\
        \hspace{1em} + CFG & $0.9487$ & $0.9604$ \\
        \hspace{1em} + CFG++ & $0.9473$ & $0.9604$ \\
        \midrule
        \rowcolor[gray]{0.9} \multicolumn{3}{l}{\textbf{Ours}} \\
        \hspace{1em} + CFG Interp. & $0.9598$ & $0.9635$ \\
        \hspace{1em} + CFG++ Interp. & \textbf{$0.9614$} & \textbf{$0.9664$} \\
        \bottomrule
    \end{tabular}
    \caption{Comparison of consistency metrics (\textbf{SC}: Subject Consistency, \textbf{BC}: Background Consistency) across different configurations using CFG and CFG++ in AnimateDiff. Our approach with interpolated CFG++ achieves the best performance, significantly enhancing both subject and background consistency.}
    \label{tab:ablation_cfg}
\end{table}

\noindent We provide quantitative analysis for using CFG and CFG++ across the Base Model, Base Model + FreeInit, and Base Model + VideoGuide (Ours) during the interpolation. 
As shown in Tab.~\ref{tab:ablation_cfg}, metrics for Base and FreeInit decrease when CFG++ is used, and metrics improve only when CFG++ is applied to our interpolation scheme. This implies the significant positive impact on consistency of CFG++ within the proposed interpolation scheme, especially compared to CFG. Also, this supports the idea that smooth interpolation of denoised samples positively impacts model performance, as discussed above.

\section{Pseudo Code}
\label{suppl:pseudocode}
\noindent Pseudo codes regarding our algorithm are provided. For clarity, the pseudo code describing our algorithm adopts the CFG++ reverse sampling framework for the entire process.

\begin{algorithm*}
\caption{VideoGuide with Sampling Diffusion Model}\label{alg:video-guide-self}
\begin{algorithmic}[1]
\REQUIRE guidance scale $\lambda \in [0, 1]$, guiding steps $I$, interpolation scale $\beta$, extra step $\tau$
\STATE Initialize $\z_T \sim \mathcal{N}(0, \rmI)$ 
\FOR{$t = T, \dots, 1$}
    \STATE $\hat{\epsilonb}_\theta(\z_t, t) = \epsilonb_\theta(\z_t, t, \phi) + \lambda[\epsilonb_\theta(\z_t, t, c) - \epsilonb_\theta(\z_t, t, \phi)]$
    \STATE $\z_{0|t}=(\z_t - \sqrt{1-\bar{\alpha}_t}\hat{\epsilonb_\theta}(\z_t, t))/\sqrt{\bar{\alpha}_t}$
    \STATE $\z_t = \sqrt{\bar{\alpha}_t}\z_{0|t} + \sqrt{1-\bar{\alpha}_t}\epsilonb, \quad \textit{where} \quad \epsilonb \sim N(0,\rmI)$
    \IF{$T - t < I$}
        \FOR{$j=0, \dots, \tau$}
            \STATE $\z_{t-j-1} = \sqrt{\bar{\alpha}_{t-j-1}}\z_{0|t-j} + \sqrt{1-\bar{\alpha}_{t-j-1}}\epsilonb_\theta(\z_{t-j}, t-j, \phi)$
        \ENDFOR
    \STATE $\z'_{0|t} = \beta \cdot \z_{0|t} + (1 - \beta) \cdot \z_{0|t-\tau}$
    \STATE $\z_{t-1} = \sqrt{\bar{\alpha}_{t-1}}\z'_{0|t} + \sqrt{1-\bar{\alpha}_{t-1}}\epsilonb_\theta(\z_t, t, \phi)$
    \STATE $\z_{t-1} = \textit{LPF}_{\gamma}(\z_{t-1}) + \textit{HPF}_{\gamma}(\epsilonb), \quad \textit{where} \quad \epsilonb \sim N(0,\rmI)$
    \ELSE
        \STATE $\z_{t-1} = \sqrt{\bar{\alpha}_{t-1}}\z_{0|t} + \sqrt{1-\bar{\alpha}_{t-1}}\epsilonb_\theta(\z_t, t, \phi)$
    \ENDIF
\ENDFOR
\STATE \textbf{Output:} Final video $\z_0$
\end{algorithmic}
\end{algorithm*}

\begin{algorithm*}
\caption{VideoGuide with Guiding Diffusion Model}\label{alg:video-guide-vc}
\begin{algorithmic}[1]
\REQUIRE guidance scale $\lambda \in [0, 1]$, guiding steps $I$, interpolation scale $\beta$, extra step $\tau$, Guiding Model $G$ parameterized by $\psi$, noise schedule $\bar{\alpha}^{(G)}$ of $G$
\STATE Initialize $\z_T \sim \mathcal{N}(0, \rmI)$ 
\FOR{$t = T, \dots, 1$}
    \STATE $\hat{\epsilonb}_\theta(\z_t, t) = \epsilonb_\theta(\z_t, t, \phi) + \lambda[\epsilonb_\theta(\z_t, t, c) - \epsilonb_\theta(\z_t, t, \phi)]$
    \STATE $\z_{0|t}=(\z_t - \sqrt{1-\bar{\alpha}_t}\hat{\epsilonb_\theta}(\z_t, t))/\sqrt{\bar{\alpha}_t}$
    \STATE $\z^{(G)}_{t} = \sqrt{\bar{\alpha}^{(G)}_t}z_{0|t} + \sqrt{1-\bar{\alpha}^{(G)}_t}\epsilonb, \quad \textit{where} \quad \epsilonb \sim N(0,\rmI) $
    \IF{$T - t < I$}
        \FOR{$j=0, \dots, \tau$}
            \STATE $\z_{0|t-j}^{(G)} = (\z_{t-j}^{(G)} - \sqrt{1-\bar{\alpha}_{t-j}^{(G)}}\hat{\epsilonb}_\psi(\z_{t-j}^{(G)}, t-j)/\sqrt{\bar{\alpha}_{t-j}^{(G)}}$
            \STATE $\z_{t-j-1}^{(G)} = \sqrt{\bar{\alpha}_{t-j-1}^{(G)}}\z_{0|t-j}^{(G)} + \sqrt{1-\bar{\alpha}_{t-j-1}^{(G)}}\epsilonb_\psi(\z_{t-j}^{(G)}, t-j, \phi)$
        \ENDFOR
    \STATE $\z'_{0|t} = \beta \cdot \z_{0|t} + (1 - \beta) \cdot \z_{0|t-\tau}^{(G)}$
    \STATE $\z_{t-1} = \sqrt{\bar{\alpha}_{t-1}}\z'_{0|t} + \sqrt{1-\bar{\alpha}_{t-1}}\epsilonb_\theta(\z_t, t, \phi)$
    \STATE $\z_{t-1} = \textit{LPF}_{\gamma}(\z_{t-1}) + \textit{HPF}_{\gamma}(\epsilonb), \quad where \quad \epsilonb \sim N(0,\rmI)$
    \ELSE
        \STATE $\z_{t-1} = \sqrt{\bar{\alpha}_{t-1}}\z_{0|t} + \sqrt{1-\bar{\alpha}_{t-1}}\epsilonb_\theta(\z_t, t, \phi)$
    \ENDIF
\ENDFOR
\STATE \textbf{Output:} Final video $\z_0$
\end{algorithmic}
\end{algorithm*}

\section{Experimental Details}
\label{suppl:experimental-details}

\subsection{Prompt Selection} \noindent
In all experiments, we utilize 800 prompts from various categories in VBench~\citep{huang2023vbench} to evaluate the model's ability to generate across diverse categories.

\subsection{Hyperparameter Selection}

\noindent We employ a classifier-free guidance (CFG) scale of $7.5$ during inference for both base models (AnimateDiff, LaVie) and FreeInit-applied cases. During interpolation of the denoised samples, we apply CFG++ reverse sampling with a guidance scale of $w = 0.8$ in DDIM $50$-step sampling. After completing the interpolation step, we revert to CFG reverse sampling with a CFG scale of $7.5$. In FreeInit, we use a Butterworth filter with a normalized frequency of $0.25$, filter order $n = 4$, and perform $5$ iterations, as recommended in prior work. The same filter is applied in our experiments with FreeInit. For AnimateDiff, we configure the guiding model with parameters $I = 5$, $\beta=0.5$, and $\tau=10$. In the case of LaVie, we set $I = 3$, $\beta = 0.5$, and $\tau = 10$ to optimize inference speed. Additionally, the $\tau$ intervals are not uniformly spaced as in the standard $50$-step DDIM sampling. To better leverage temporally consistent samples, we divide the remaining interval into $25$ steps for reverse sampling during guidance steps.


\subsection{Figure Explanation} \noindent
\textbf{Base models used for Figure \ref{fig:qualitative}}: \\
(a) AnimateDiff with pretrained T2I model RealisticVision. \\
(b) AnimateDiff with pretrained T2I model RealisticVision. \\
(c) LaVie. \\
(d) LaVie. \\
\textbf{Base model used for Figure \ref{fig:prior_distillation}}: \\
AnimateDiff with pretrained T2I model ToonYou.

\section{User Study}
\label{suppl:user-study}

\noindent We conduct a user study to evaluate generated video samples using three criteria: \textbf{Text Alignment}, \textbf{Overall Quality}, and \textbf{Smooth and Dynamic Motion}, with all metrics scored on a 1 to 5 scale. A total of 30 participants provided ratings for each metric, offering comprehensive feedback on the generated videos. Tab.~\ref{tab:user_study} shows that our method surpasses the baseline and previous work in all evaluated aspects.

\noindent \textbf{Text Alignment}
\begin{itemize}
   \item Measures how well the video corresponds to the prompt, focusing on semantic coherence.
   \item Question: Do you think the videos reflect the given text condition well? (5: Strongly Agree / 4: Agree / 3: Neutral / 2: Disagree / 1: Strongly Disagree)

\end{itemize}

\noindent \textbf{Overall Quality}
\begin{itemize}
   \item Assesses the video's visual consistency, image degradation, and aesthetic appeal.
   \item Question: Do you think the video's overall quality is good? (rich detail, unchanging objects) (5: Strongly Agree / 4: Agree / 3: Neutral / 2: Disagree / 1: Strongly Disagree)
   
\end{itemize}

\noindent \textbf{Smooth and Dynamic Motion}
\begin{itemize}
   \item Evaluates the naturalness and fluidity of the motion in the video.
   \item Question: Do you think the video's overall motion is smooth and dynamic? (5: Strongly Agree / 4: Agree / 3: Neutral / 2: Disagree / 1: Strongly Disagree)

\end{itemize}

\begin{table}[h]
    \centering
    \small
    \begin{tabular}{l | ccc}
        \toprule
        Method & TA & OQ & SDM \\
        \midrule
        Base    & $3.72$ & $2.84$ & $2.9$ \\
        Base + FreeInit & $\underline{3.97}$ & $\underline{3.35}$ & $\underline{3.38}$ \\
        Base + VideoGuide (Ours) & $\bm{4.36}$ & $\bm{4.37}$ & $\bm{4.36}$ \\
        \bottomrule
    \end{tabular}
    \caption{User Study. Text Alignment (TA), Overall Quality (OQ), Smooth and Dynamic Motion (SDM) are evaluated among methods. \textbf{Bold}: best, \underline{underline}: second best.}
    \label{tab:user_study}
\end{table}

\section{Application on DiT-based Models}
\label{suppl:dit-based}
\noindent We further evaluate the robustness of our methodology by applying it to different architectures and schedulers. Specifically, we present further evaluation on models that use Diffusion Transformer (DiT)~\citep{Peebles2022DiT} architecture: Open-Sora v1.0~\citep{opensora} and Open-Sora v1.2~\citep{opensora}. Each model employs a standard DDIM scheduler (50 steps) and a rectified flow~\citep{liu2022flow} scheduler, respectively. In the rectified flow-based configuration, the objective for training can be formulated as follows:
\begin{equation}
    \begin{aligned}
        \z_t &= (1-t)\z_0 + t\epsilonb \quad where \quad t \in [0,1] \\ 
        \hat{\theta} &= \underset{\theta}{\mathrm{argmin}}\, \mathbb{E}\left[||(\z_0 - \epsilonb) - v_\theta(\z_t, t)||^2_2\right]  
    \end{aligned}
\end{equation}
Using the objective above we can redefine our method as below:
\begin{equation}
    \begin{aligned}
        \z_{0|t_i} & = \z_{t_i} + t_i \cdot v_\theta(\z_{t_i}, t_i) \\
        \epsilonb_\theta(\z_{t_i},t_i) &= \z_{t_i} - (1-t_i) \cdot v_\theta(\z_{t_i}, t_i) \\ 
        \z_{t_{i-1}} &= (1-t_{i-1})f(\z_{0|t_i}, \beta, \tau)+t_{i-1}\epsilonb_\theta(\z_{t_i}, t_i)
    \end{aligned}
\end{equation}
where $f(\z_{0|t_i}, \beta, \tau)$ is the interpolation function between $\z_{0|t_i}$ and $\z_{0|t_{i-\tau}}$ with scale $\beta$. The results in Tab.~\ref{tab:results_dit} demonstrate that our method improves temporal consistency for both baselines while preserving imaging quality and introducing only a minimal decrease in dynamic degree. These findings indicate that our methodology enhances performance regardless of the underlying architecture and scheduler.

\begin{table*}[t]
\centering
\resizebox{\textwidth}{!}{%
\fontsize{6}{8}\selectfont
\begin{tabularx}{\textwidth}{l>{\centering\arraybackslash}X>{\centering\arraybackslash}X>{\centering\arraybackslash}X>{\centering\arraybackslash}X>{\centering\arraybackslash}X} 
\toprule
\textbf{Method} & \makecell{\textbf{Subject} \\ \raggedright\textbf{consistency} ($\uparrow$)} & \makecell{\textbf{Background} \\ \textbf{Consistency} ($\uparrow$)} & \makecell{\textbf{Imaging} \\ \textbf{Quality ($\uparrow$)}} &  \makecell{\textbf{Motion} \\ \textbf{Smoothness} ($\uparrow$)} & \makecell{\textbf{Dynamic} \\ \textbf{Degree} ($\uparrow$)} \\

\cmidrule{1-6}
OpenSora v1.0~\citep{opensora} (DDIM~\citep{song2020denoising}) & $0.9735$ & $0.9689$ & $0.6615$ & $0.9678$ & $4.97$ \\
OpenSora v1.0 + VideoGuide (self-guided) & $0.9763$ & $0.9689$ & $0.6738$ & $0.9754$ & $3.88$ \\
OpenSora v1.2~\citep{opensora} (Rectified Flow~\citep{liu2022flow}) & $0.9725$ & $0.9696$ & $0.6582$ & $0.9881$ & $12.68$ \\
OpenSora v1.2 + VideoGuide (self-guided) & $0.9808$ & $0.9748$ & $0.6689$ & $0.9903$ & $11.07$ \\

\bottomrule
\end{tabularx}
}
\caption{Quantitative comparison of video generation in DiT-based architecture.}
\label{tab:results_dit}
\end{table*}

\begin{figure*}[b]
    \centering
    \includegraphics[width=\textwidth]{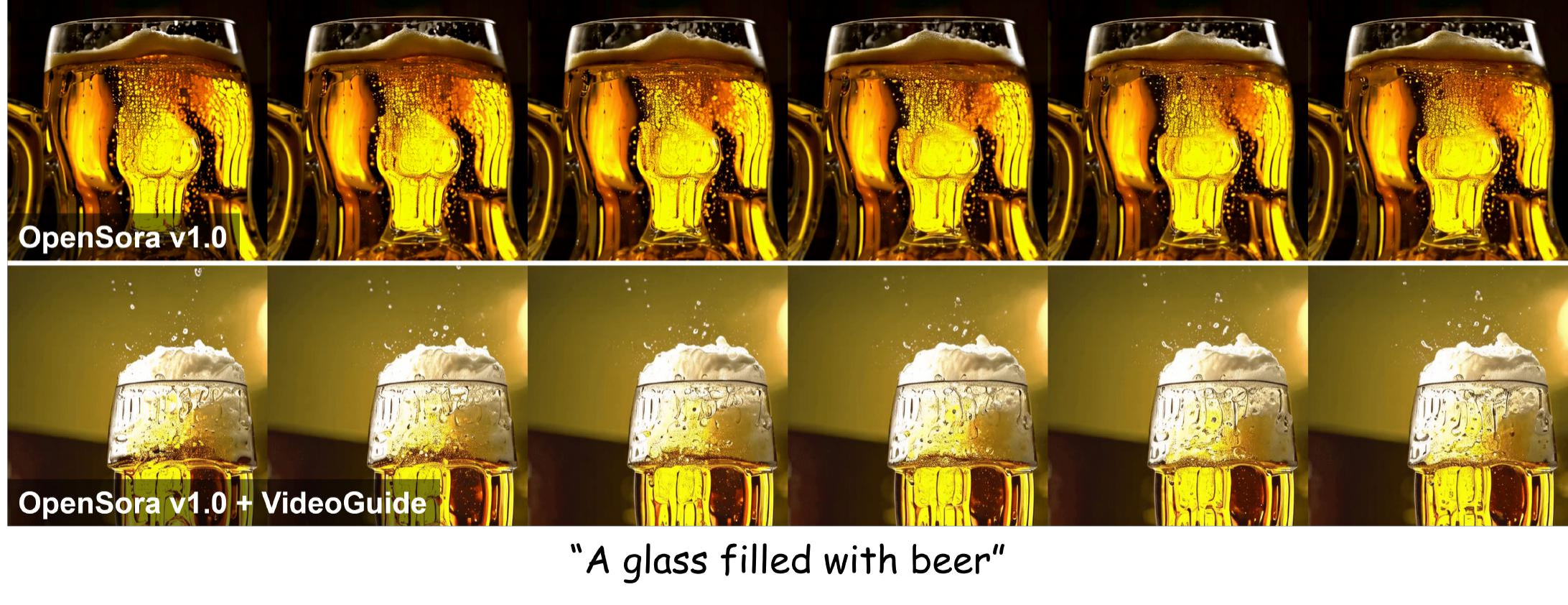}
    \caption{Qualitative Results of VideoGuide on Open-Sora v1.0.}
\end{figure*}

\begin{figure*}[b]
    \centering
    \includegraphics[width=\textwidth]{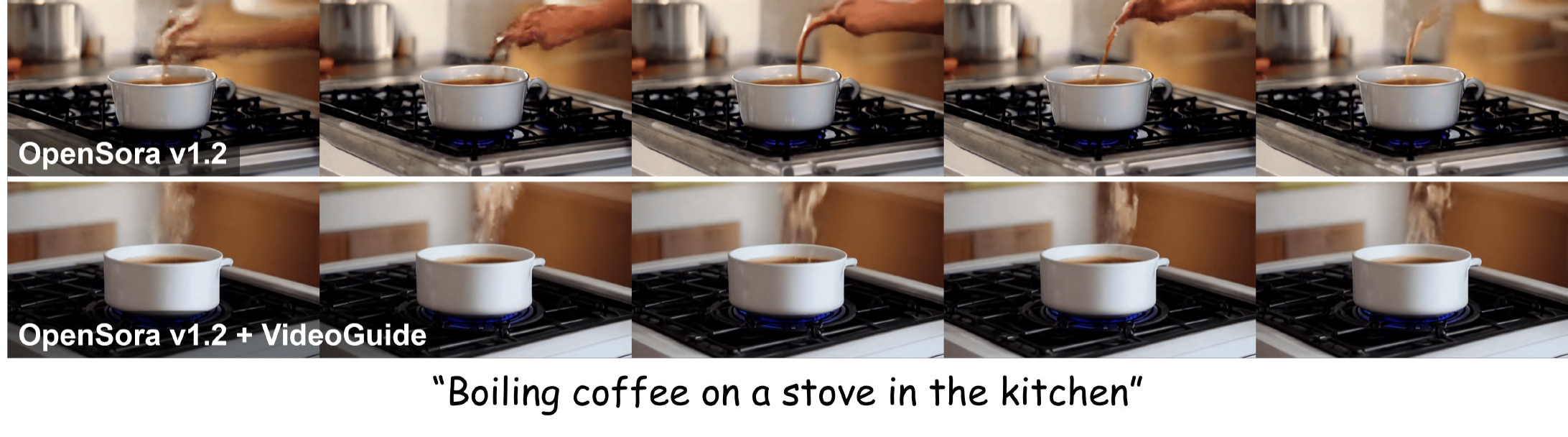}
    \caption{Qualitative Results of VideoGuide on Open-Sora v1.2.}
\end{figure*}

\section{Comparison with Orthogonal Methods}
\label{suppl:orthogonal}

\noindent A recent study, UniCtrl~\citep{chen2024unictrl}, attempts to improve semantic consistency and motion quality in an  approach orthogonal to ours. In this section, we compare the performance of each technique and assess the feasibility of combining them. Following the recommendation in the paper, we use a motion injection degree of $c=0.2$, while maintaining the same experimental configuration as described in Section~\ref{suppl:experimental-details}.
As illustrated in Table~\ref{tab:results_ortho}, UniCtrl~\citep{chen2024unictrl} improves temporal consistency but at the cost of a significant reduction in dynamic degree and imaging quality.

\begin{table*}[t]
\centering
\resizebox{\textwidth}{!}{%
\fontsize{6}{8}\selectfont
\begin{tabularx}{\textwidth}{l>{\centering\arraybackslash}X>{\centering\arraybackslash}X>{\centering\arraybackslash}X>{\centering\arraybackslash}X>{\centering\arraybackslash}X} 
\toprule
\textbf{Method} & \makecell{\textbf{Subject} \\ \raggedright\textbf{consistency} ($\uparrow$)} & \makecell{\textbf{Background} \\ \textbf{Consistency} ($\uparrow$)} & \makecell{\textbf{Imaging} \\ \textbf{Quality ($\uparrow$)}} &  \makecell{\textbf{Motion} \\ \textbf{Smoothness} ($\uparrow$)} & \makecell{\textbf{Dynamic} \\ \textbf{Degree} ($\uparrow$)} \\

\cmidrule{1-6}
AnimateDiff~\citep{guo2023animatediff} & $0.9183$ & $0.9437$ & $0.6647$ & $0.9547$ & $26.67$ \\
AnimateDiff + UniCtrl~\citep{chen2024unictrl} & $0.9259$ & $0.9413$ & $0.6032$ & $0.9584$ & $14.96$ \\
AnimateDiff + Ours & $0.9614$ & $0.9664$ & $0.6671$ & $0.9772$ & $16.78$ \\
AnimateDiff + UniCtrl + Ours & $0.9639$ & $0.9628$ & $0.5883$& $0.9776$ & $5.02$ \\

\bottomrule
\end{tabularx}
}
\caption{Quantitative comparison with orthogonal methods.}
\label{tab:results_ortho}
\end{table*}

\begin{figure*}
    \centering
    \includegraphics[width=\textwidth]{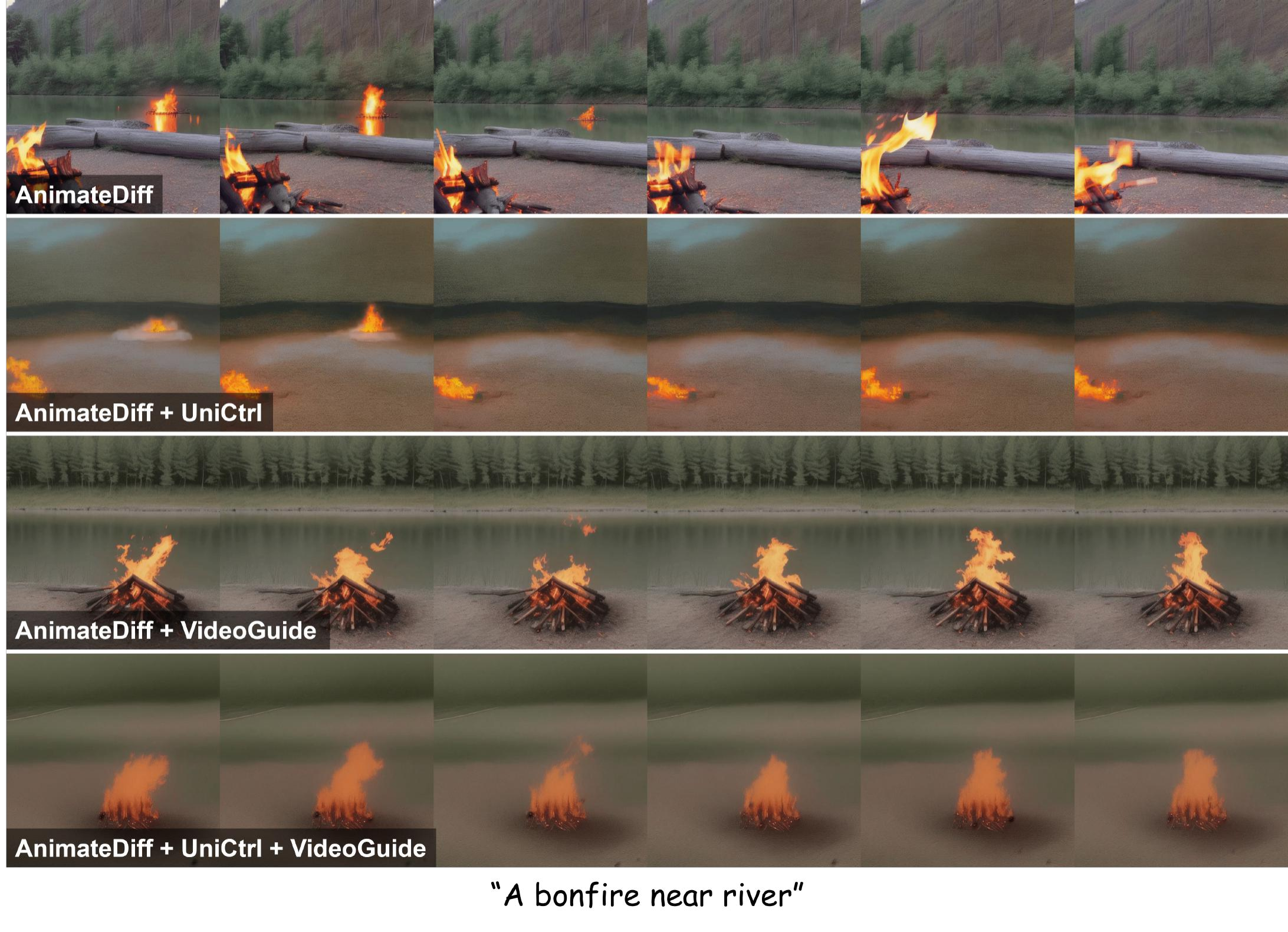}
    \caption{Qualitative Comparison of UniCtrl and VideoGuide.}
\end{figure*}

\section{Limitations}
\label{suppl:limitations}
\noindent While our approach significantly improves the performance of baseline models, it relies on sharing the same Variational Auto-Encoder(VAE)~\citep{kingma2013auto} space. In practice, many latent diffusion models utilize the same VAE, making this requirement generally feasible. However, if the VAE spaces differ, one potential solution is to decode, interpolate, and re-encode the features. This process, however, incurs additional computational overhead and risks losing fine details due to iterative encoding-decoding. Developing an effective method to address compatibility across different VAE spaces remains an avenue for future research.

\section{More Qualitative Examples} \noindent
Additional samples are provided in following pages:
\begin{itemize}
   \item Supplemental examples of prior distillation.
   \item Qualitative comparison for various methods.
   \item Qualitative comparison for various base models.
   \item Usage of VideoGuide to solve sudden frame shifts in LaVie samples.
\end{itemize}

\label{suppl:qualitative}
\subsection{Prior Distillation}
\begin{figure*}
    \centering
    \includegraphics[width=0.9\linewidth]{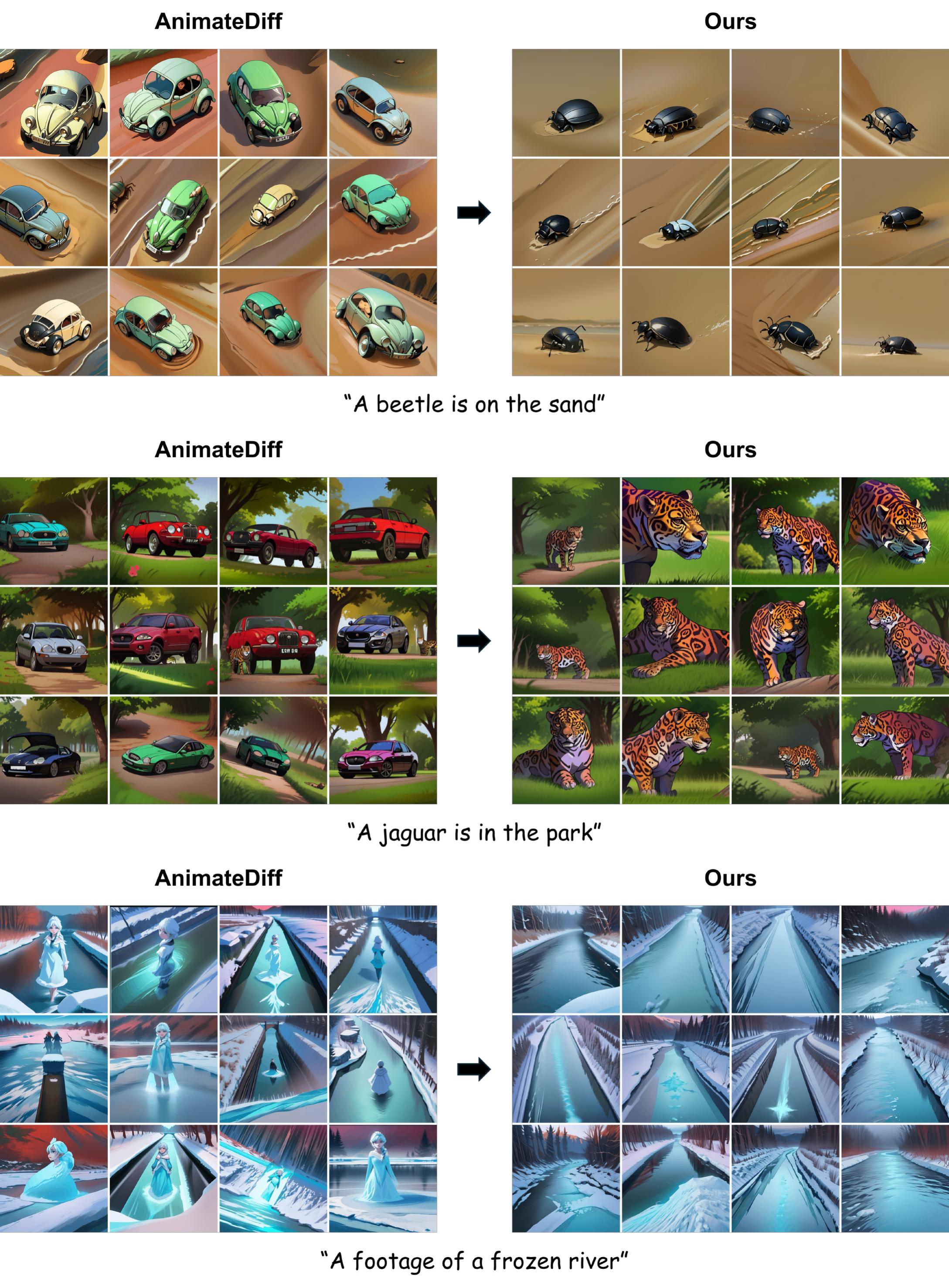}
    \caption{\textbf{Prior Distillation}. For each prompt, we share the same random seed for both methods.}
\end{figure*}

\subsection{More Qualitative Comparison Results}
\begin{figure*}
    \centering
    \includegraphics[width=0.8\linewidth]{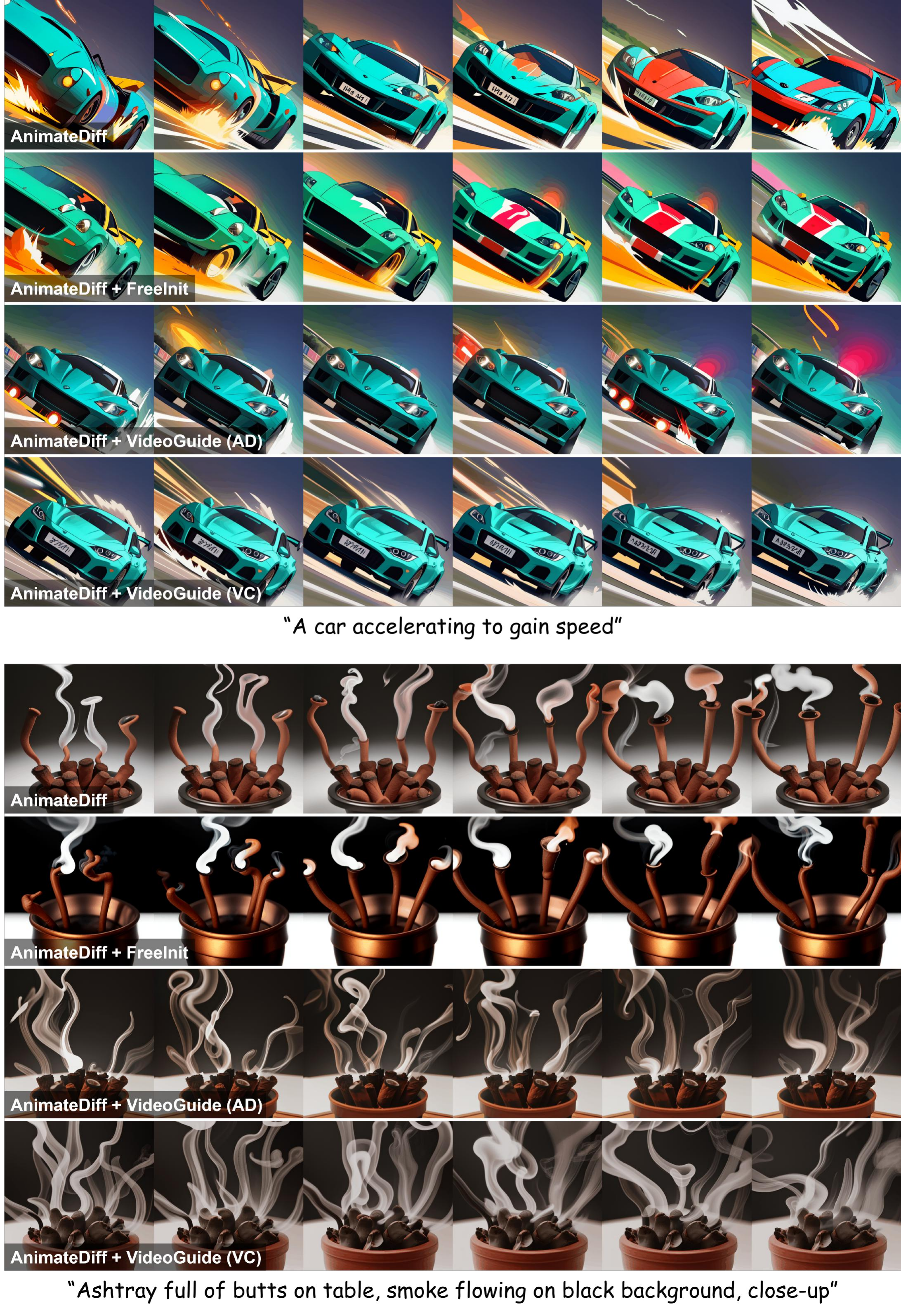}
    \caption{More Qualitative Comparison Results of VideoGuide. Top: AnimateDiff with ToonYou, Bottom: AnimateDiff with RCNZCartoon}
\end{figure*}

\subsection{More Qualitative Results}
\begin{figure*}
    \centering
    \includegraphics[width=0.8\linewidth]{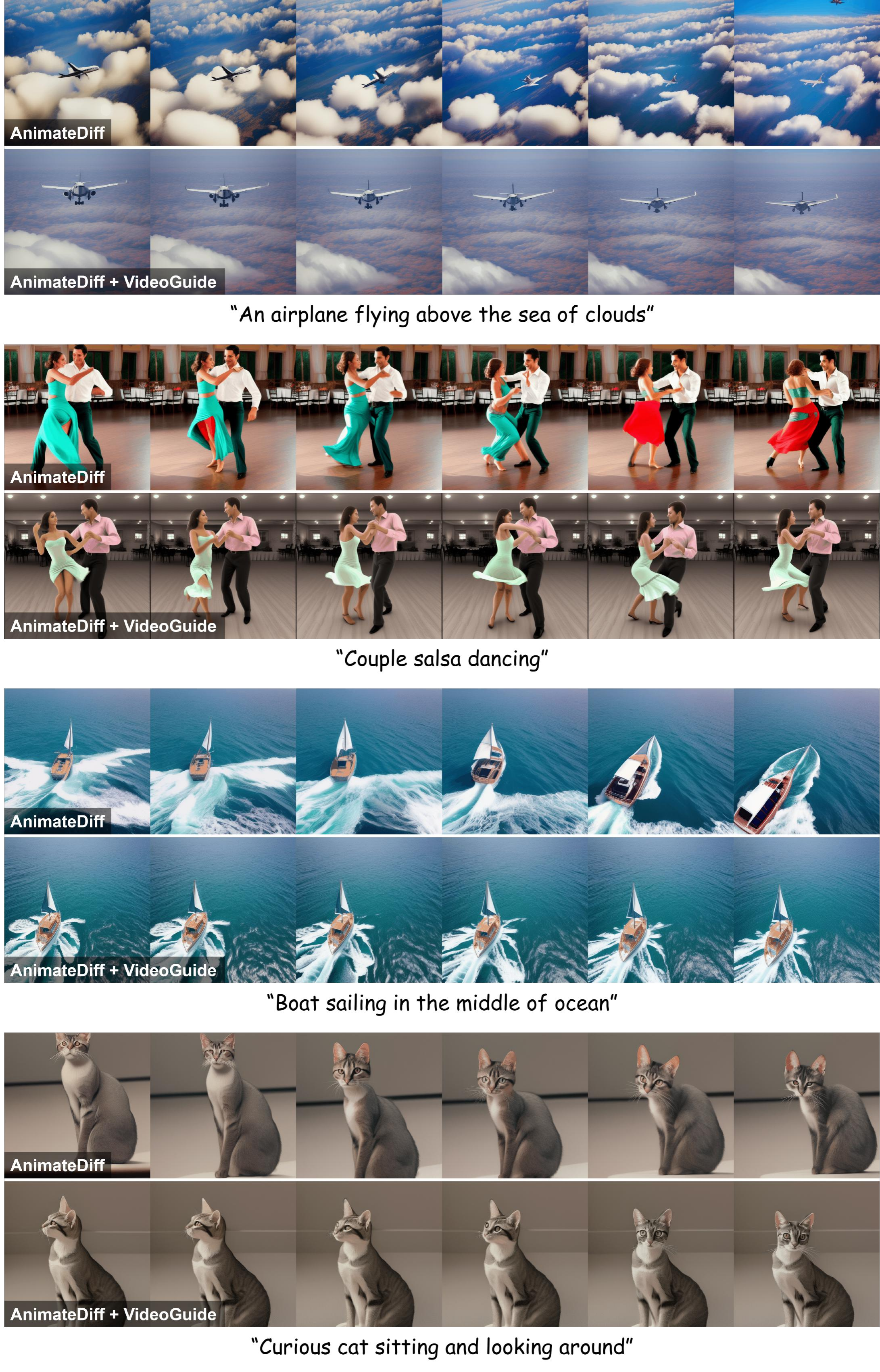}
    \caption{More Qualitative Results of VideoGuide on AnimateDiff (with RealisticVision).}
\end{figure*}

\begin{figure*}
    \centering
    \includegraphics[width=0.8\linewidth]{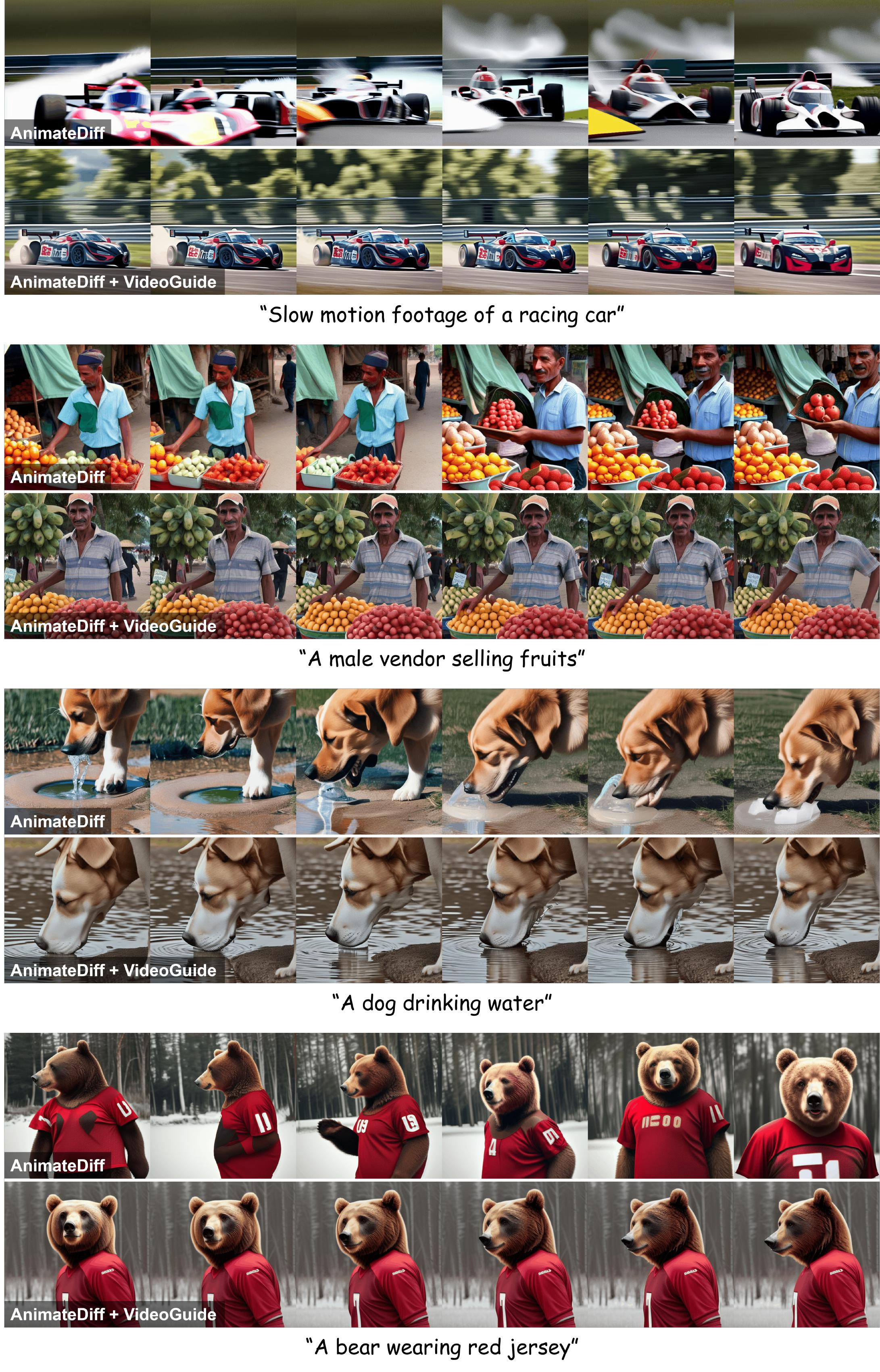}
    \caption{More Qualitative Results of VideoGuide on AnimateDiff (with RealisticVision).}
\end{figure*}

\begin{figure*}
    \centering
    \includegraphics[width=0.8\linewidth]{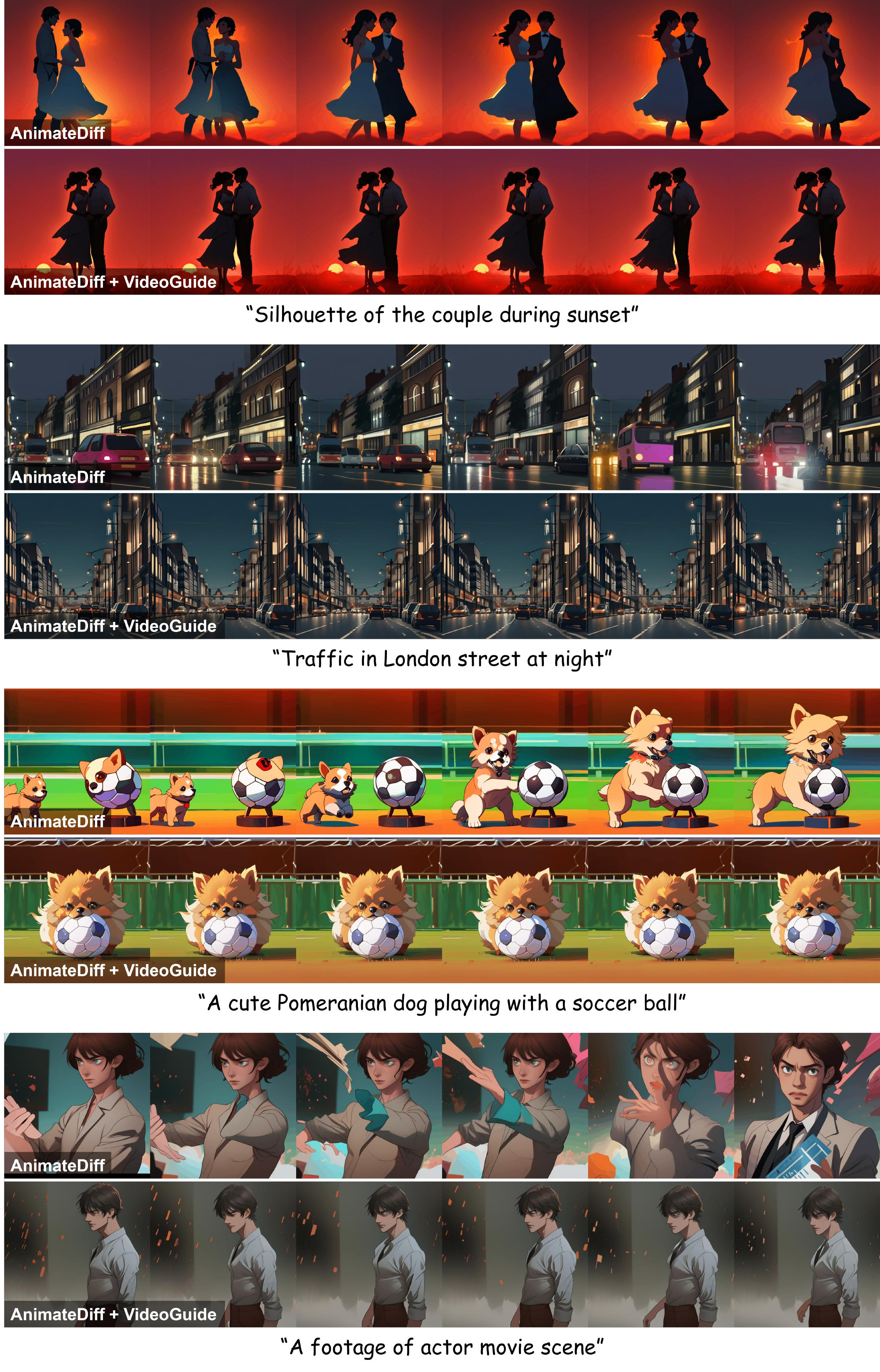}
    \caption{More Qualitative Results of VideoGuide on AnimateDiff (with ToonYou).}
\end{figure*}

\begin{figure*}
    \centering
    \includegraphics[width=0.8\linewidth]{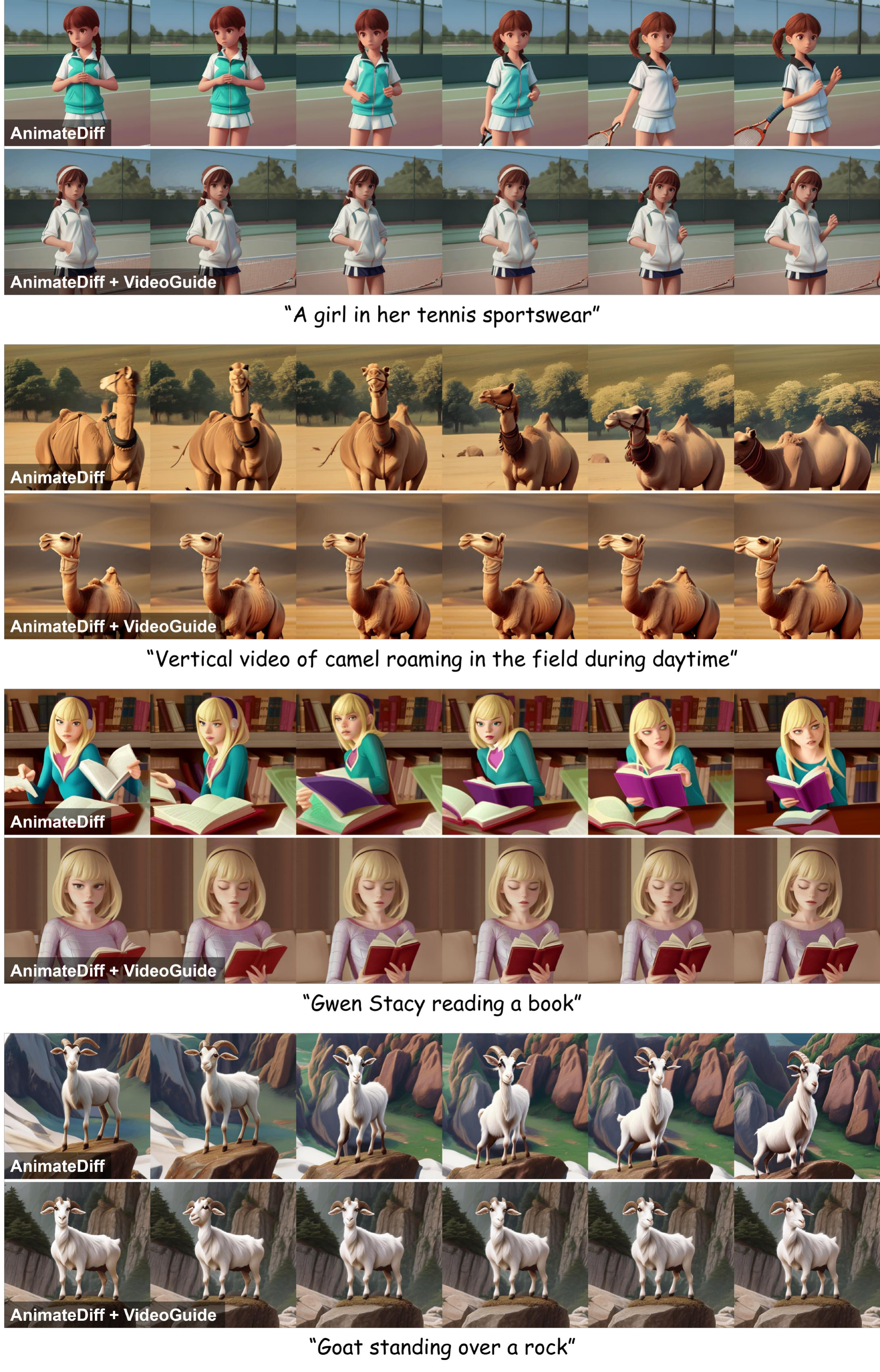}
    \caption{More Qualitative Results of VideoGuide on AnimateDiff (with RCNZCartoon).}
\end{figure*}

\begin{figure*}
    \centering
    \includegraphics[width=0.8\linewidth]{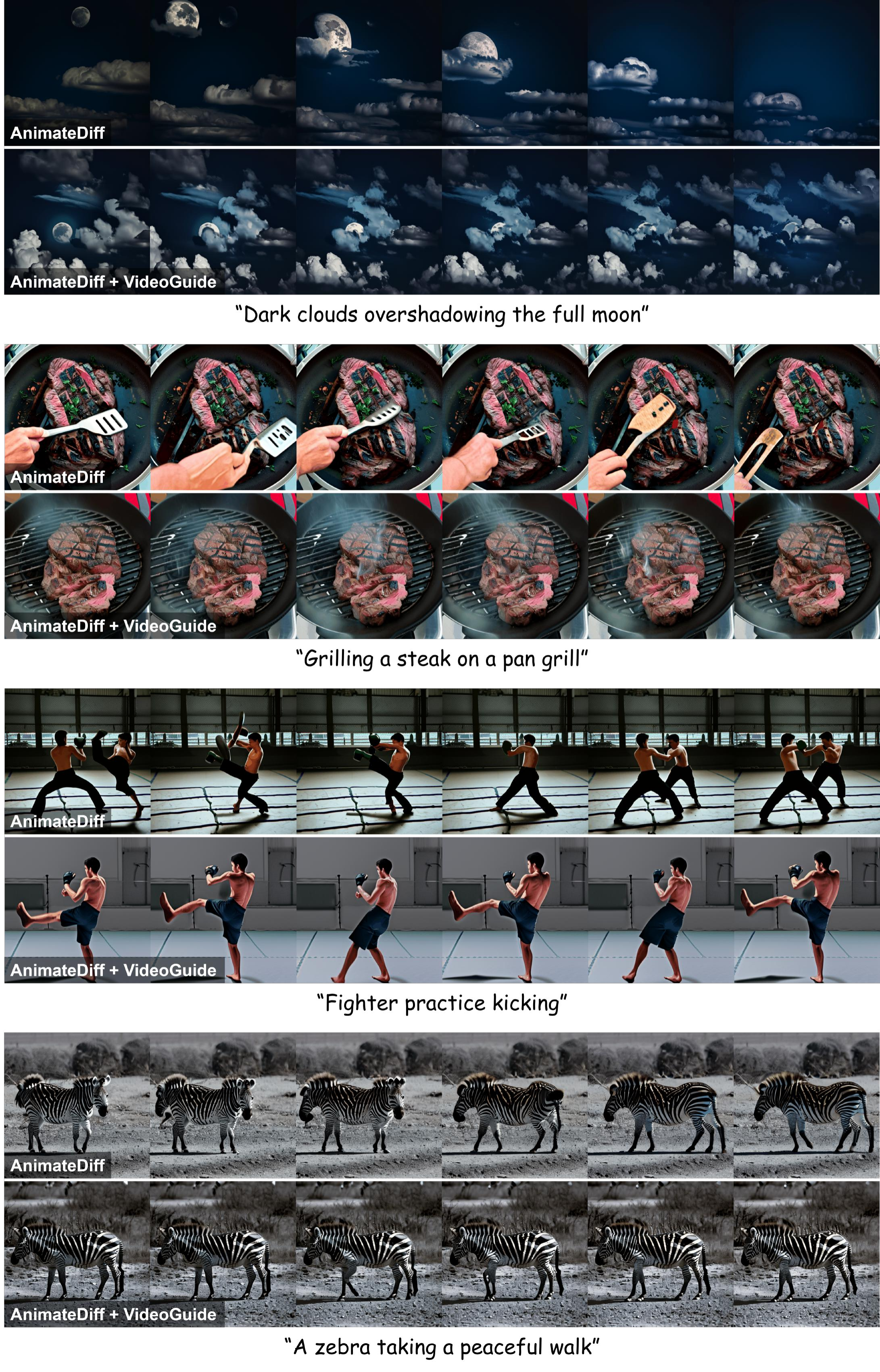}
    \caption{More Qualitative Results of VideoGuide on AnimateDiff (with FilmVelvia).}
\end{figure*}

\begin{figure*}
    \centering
    \includegraphics[width=0.9\linewidth]{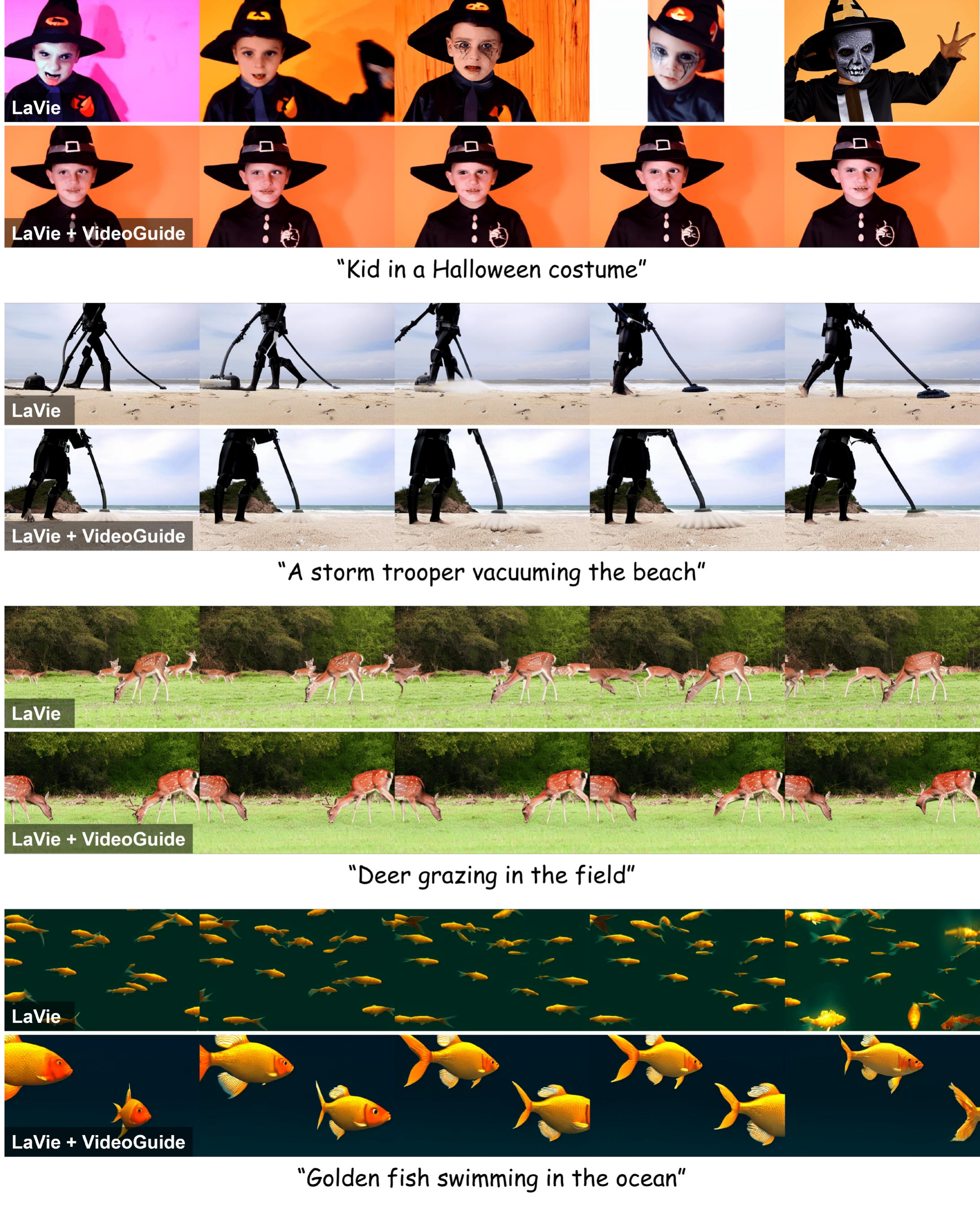}
    \caption{More Qualitative Results of VideoGuide on LaVie.}
\end{figure*}

\begin{figure*}
    \centering
    \includegraphics[width=0.9\linewidth]{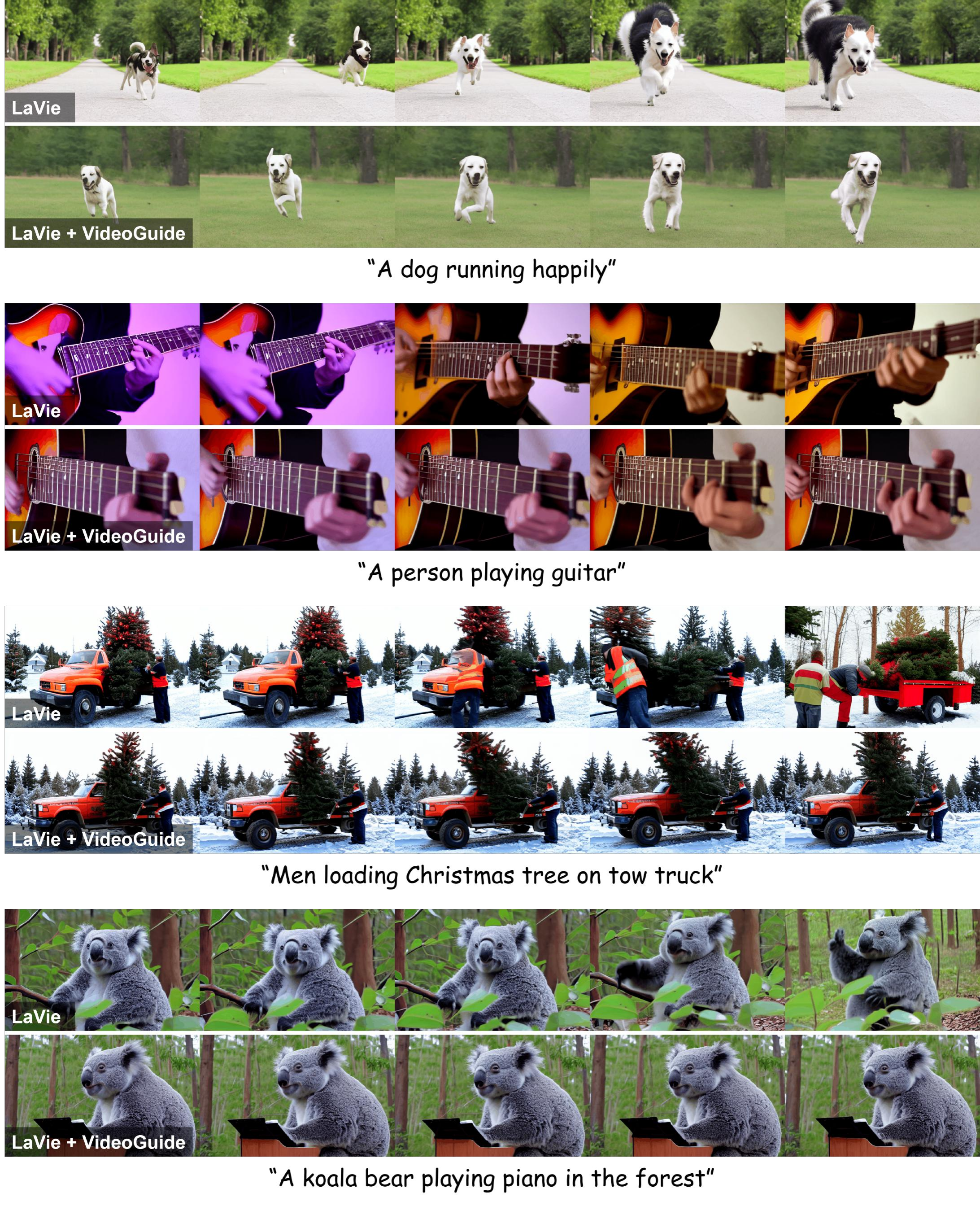}
    \caption{More Qualitative Results of VideoGuide on LaVie.}
\end{figure*}

\subsection{LaVie Sudden Shift}
\begin{figure*}
    \centering
    \includegraphics[width=0.9\linewidth]{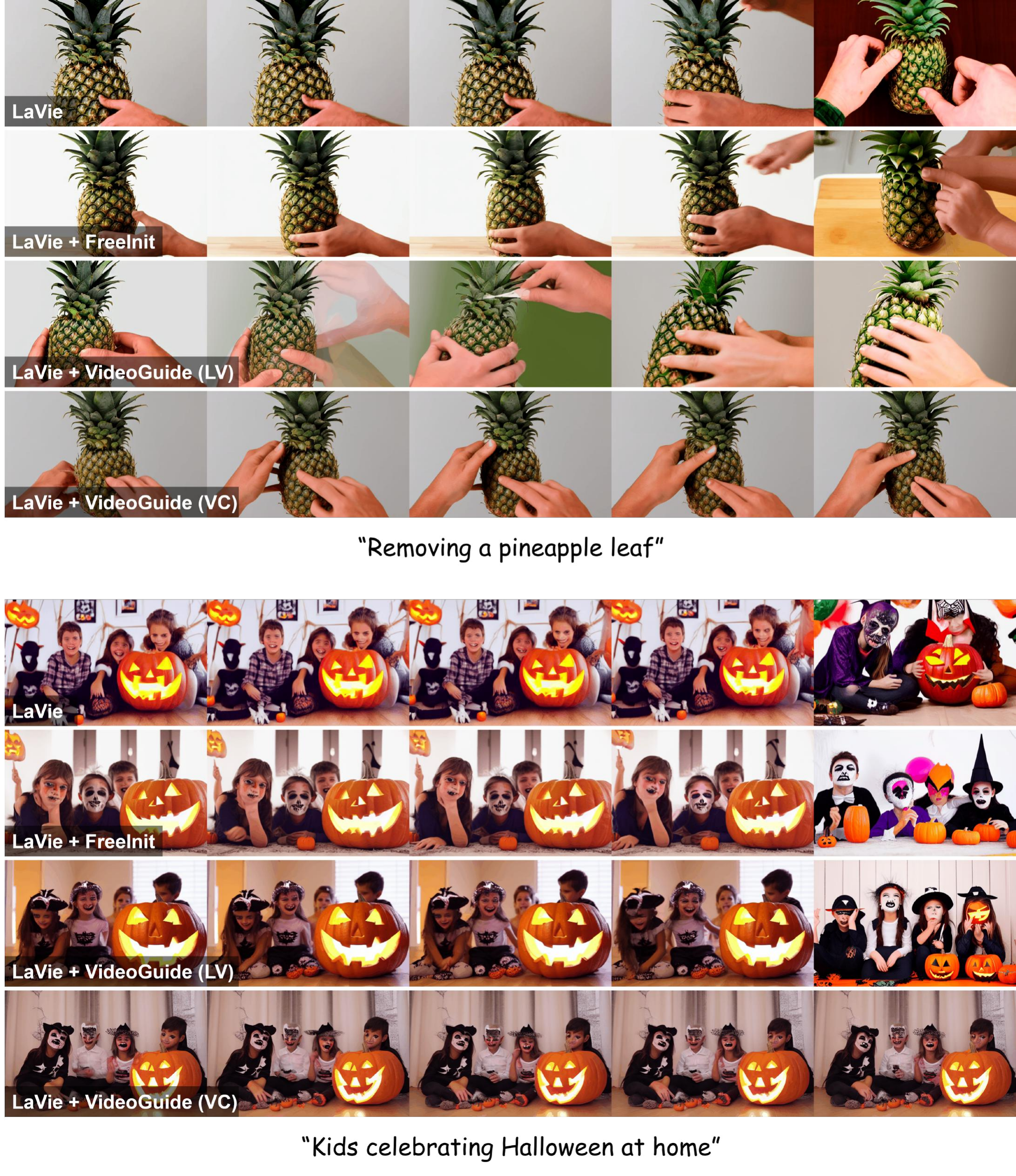}
    \caption{VideoGuide helps solve the issue of sudden frame shifts in LaVie samples. By integrating an external guiding model, VideoGuide provides smoother frame transitions to the base model. LV indicates that guidance model of LaVie is used (the self-guided case), and VC indicates that guidance model of VideoCrafter2 is used. Guidance given with the external model VideoCrafter2 solves sudden frame shift unsolvable by other methods.}
\end{figure*}

\end{document}